\documentclass[10pt,journal,compsoc]{IEEEtran}
\ifCLASSOPTIONcompsoc
  \usepackage[nocompress]{cite}
\else
  \usepackage{cite}
\fi
\ifCLASSINFOpdf
   \usepackage[pdftex]{graphicx}
   \graphicspath{{figures/}}
   \DeclareGraphicsExtensions{.pdf,.jpeg,.png}
\else
\fi
\usepackage{amsmath}
\usepackage{algorithmic}
\usepackage{array}
\usepackage{bbm}
\usepackage{url}
\usepackage{multirow}
\usepackage{makecell}
\usepackage{amssymb}
\usepackage{xcolor}
\usepackage{float}

\newcommand{\etal}{\textit{et al.}}
\ifCLASSOPTIONcompsoc
 \usepackage[caption=false,font=footnotesize,labelfont=sf,textfont=sf]{subfig}
\else
 \usepackage[caption=false,font=footnotesize]{subfig}
\fi
\def\Equal{\texttt{=}}

\begin{document}
\title{A Unified Visual Information Preservation Framework for Self-supervised Pre-training in Medical Image Analysis}

\author{Hong-Yu~Zhou,~\IEEEmembership{Student Member,~IEEE},
        Chixiang~Lu,
        Chaoqi~Chen,
        Sibei~Yang,
        and~Yizhou~Yu,~\IEEEmembership{Fellow,~IEEE,}
\IEEEcompsocitemizethanks{\IEEEcompsocthanksitem Hong-Yu Zhou, Chixiang Lu, Chaoqi Chen, and Yizhou Yu are with the Department of Computer Science, The University of Hong Kong, Hong Kong. Email:
\{whuzhouhongyu, luchixiang, cqchen1994\}@gmail.com, yizhouy@acm.org.\protect
\IEEEcompsocthanksitem Sibei Yang is with ShanghaiTech University and Shanghai Engineering Research Center of Intelligent Vision and Imaging, Shanghai, China. Email: yangsb@shanghaitech.edu.cn.\protect
\IEEEcompsocthanksitem First two authors contributed equally.\protect
\IEEEcompsocthanksitem Corresponding author: Sibei Yang and Yizhou Yu.
}
}


\markboth{}%
{Shell \MakeLowercase{\textit{et al.}}: Bare Demo of IEEEtran.cls for Computer Society Journals}



\IEEEtitleabstractindextext{%
\begin{abstract}
Recent advances in self-supervised learning (SSL) in computer vision are primarily comparative, whose goal is to preserve invariant and discriminative semantics in latent representations by comparing siamese image views. However, the preserved high-level semantics do not contain enough local information, which is vital in medical image analysis (e.g., image-based diagnosis and tumor segmentation). To mitigate the locality problem of comparative SSL, we propose to incorporate the task of pixel restoration for explicitly encoding more pixel-level information into high-level semantics. We also address the preservation of scale information, a powerful tool in aiding image understanding but has not drawn much attention in SSL. The resulting framework can be formulated as a multi-task optimization problem on the feature pyramid. Specifically, we conduct multi-scale pixel restoration and siamese feature comparison in the pyramid. In addition, we propose non-skip U-Net to build the feature pyramid and develop sub-crop to replace multi-crop in 3D medical imaging. The proposed unified SSL framework (PCRLv2) surpasses its self-supervised counterparts on various tasks, including brain tumor segmentation (BraTS 2018), chest pathology identification (ChestX-ray, CheXpert), pulmonary nodule detection (LUNA), and abdominal organ segmentation (LiTS), sometimes outperforming them by large margins with limited annotations. Codes and models are available at \url{https://github.com/RL4M/PCRLv2}.
\end{abstract}

}

\maketitle

\IEEEdisplaynontitleabstractindextext

% \fi
\IEEEpeerreviewmaketitle

\IEEEraisesectionheading{\section{Introduction}\label{sec:introduction}}

\IEEEPARstart{I}t is usual to acquire a substantial amount of manually labeled data before training deep neural networks. This condition is easy to meet in natural images, where labor costs and labeling difficulties are tolerable. In medical image analysis, however, credible annotations are mainly derived from domain experts' diagnoses, which are challenging to obtain due to the rarity of the target disease, the need to safeguard patient privacy, and the scarcity of medical resources. Against this background, self-supervised learning (SSL) has been widely accepted as a viable technique to learn medical image representations without specialistic annotations. We usually deploy SSL in the pre-training stage to obtain well-transferable features, which can be transferred to various downstream tasks for performance boosting.

% Existing SSL methodologies can be roughly grouped into two categories, i.e., pretext-based and comparison-based SSL, according to how they acquire supervision signals. Specifically, pretext-based approaches are built on top of a wide range of pretext tasks. By solving pre-defined tasks, SSL can extract meaningful latent representations from unlabeled data. Prior to comparative SSL, 

Recent advances in SSL are mostly based on comparative learning~\cite{he2020momentum,chen2020simple,grill2020bootstrap,chen2021exploring}. The rationale behind is to learn transferable latent representations with invariant and discriminative semantics by maximizing the mutual information between a pair of siamese images. One potential problem of these comparative methods is that they mainly focus on encoding high-level global semantics in representations but ignore the preservation of pixel-level information\footnote{In 3D medical images, we often use ``voxel'' to denote the same concept as the pixel does in 2D images. For simplicity, we use ``pixel'' to denote the smallest addressable element in both 2D and 3D images in the rest of this paper.}. However, in medical image analysis, the latter type of information usually plays a vital role. For instance, in chest pathology detection, radiologists or clinicians are required to point out small lesions from a chest X-ray according to their textures. Sometimes, these areas of pathologies are so hard to identify that even medical experts have to check pixel-level details to tell where the lesions are. Another typical example lies in brain tumor segmentation, where the segmentation error of one voxel may cause irreparable harm to patients in brain surgeries, such as a permanent damage to the cochlear nerve when trying to remove the acoustic neuroma.

An intuitive way to preserve pixel-level information in learned features is to restore the pixel-level content from latent representations directly. This methodology, known as context restoration~\cite{pathak2016context}, has already been adopted as a surrogate task in pretext-based SSL for natural~\cite{pathak2016context,larsson2016learning,zhang2017split} and medical images~\cite{chen2019self,zhou2021models}. Specifically, these approaches first apply various data augmentation strategies to a given image to generate a corrupted input, based on which deep models are trained to restore original pixels. In this way, we explicitly require the latent representations to preserve information closely related to pixels. Although pure pixel-based features are not as transferable as those from comparative SSL~\cite{he2020momentum,zhou2020comparing}, we hypothesize it is still beneficial to explicitly preserve pixel-level information and global semantics, especially in medical image analysis where details matter a lot.

\begin{figure}[t]
    \centering
    \includegraphics[width=1.0\columnwidth]{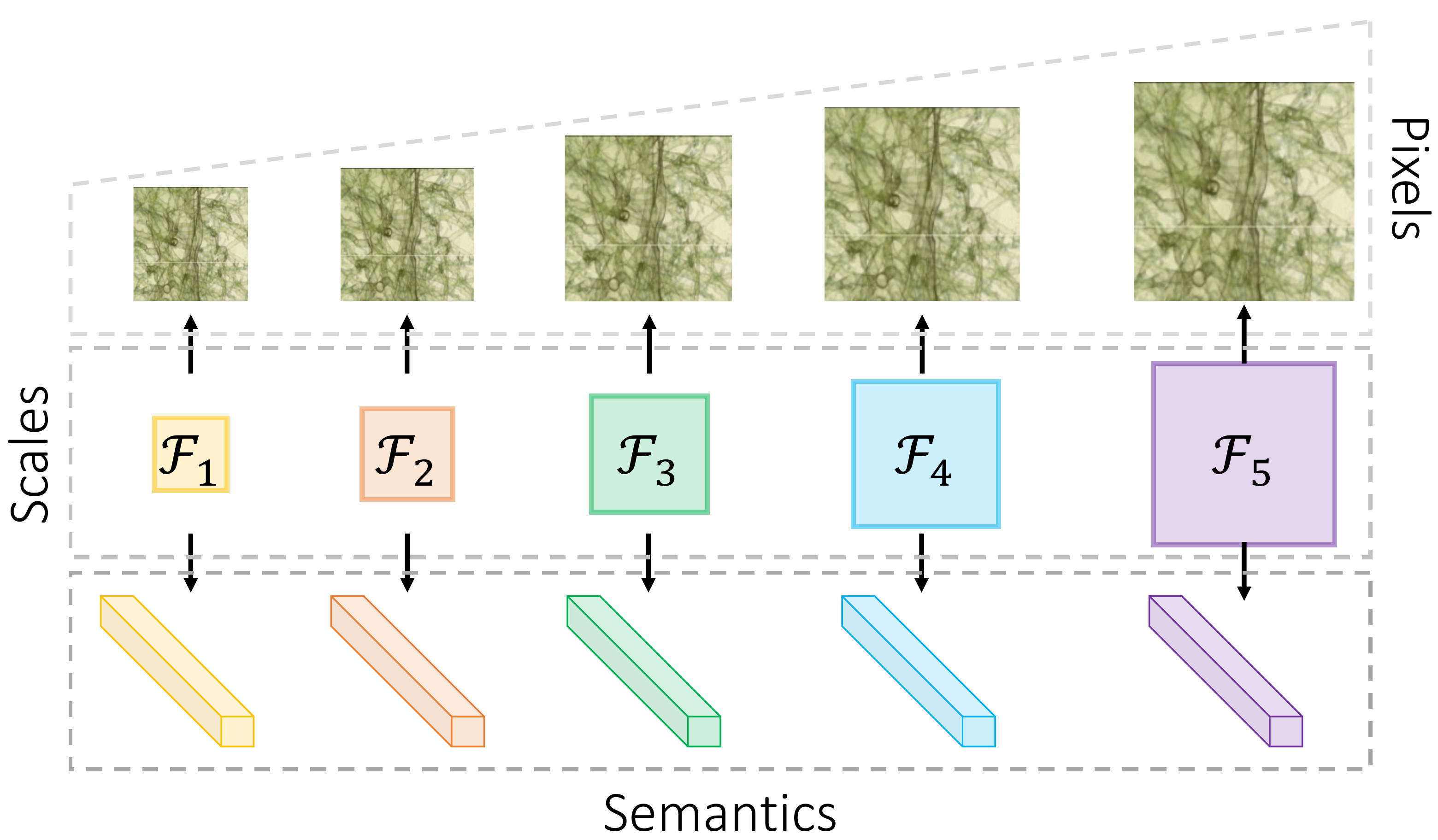}
    \caption{Motivation illustration. We propose a unified SSL framework to simultaneously preserve information in visual representations from perspectives of pixels, semantics, and scales. $\{\mathcal{F}_1, \mathcal{F}_2, \mathcal{F}_3, \mathcal{F}_4, \mathcal{F}_5\}$ denote different levels in the feature pyramid, given an input image. Our approach restores uncorrupted inputs from the feature maps directly to preserve pixel-level details. In order to retain the global semantic information, our method compares siamese one-dimensional representations. Last but not the least, the proposed methodology conducts pixel restoration and feature comparison at different scales. The rationale behind is to introduce multi-scale self-supervised latent representations, making them more transferable to various downstream tasks.}
    \label{intro}
\end{figure}

Besides semantics and pixels, introducing multi-scale representations has been proven to be quite helpful in aiding image understanding~\cite{dalal2005histograms,lowe2004distinctive,yang2015multi,lin2017feature,long2015fully,ronneberger2015u}. The common practice of these methods is to construct a feature pyramid during training, testing, or both stages. Then, various tasks, such as detection, and segmentation, can be conducted on the basis of multi-scale features. The goal of building the feature pyramid is to endow image representations with the ability to recognize objects at different scales, which is also consistent with the law of human cognition~\cite{romeny2008front}. However, the preservation of visual information at multiple scales is rarely mentioned in SSL. Thus, it is unclear whether introducing multi-scale self-supervised representations provides a stronger transfer learning ability.

In Figure~\ref{intro}, we illustrate the motivation of the proposed unified visual information preservation framework for SSL. The introduced framework addresses the preservation of information in self-supervised visual representations from three aspects: pixels, semantics, and scales. Firstly, to retain pixel-level information in latent representations, our framework involves a reconstruction branch in the self-supervised model to rebuild uncorrupted images from corrupted inputs. Specifically, we ask the self-supervised model to restore pixels from feature maps of randomly corrupted inputs during training. As a result, information closely associated with pixels can be explicitly encoded into the latent representations. In practice, this type of information would enhance the ability of self-supervised representations to recognize and differentiate textures. Apart from pixel-level information, preserving invariant and discriminative semantics in visual representations is also necessary. Towards this end, we adopt the existing comparative SSL to encode invariant semantic information by comparing high-level representations of siamese image patches~\cite{chen2021exploring}. We empirically found the siamese SSL not only produces comparably (sometimes more) transferable medical image representations but also is much easier to implement in comparison to the typical contrastive manner~\cite{he2020momentum}. Last but not the least, the proposed unified framework introduces multi-scale latent representations by conducting pixel restoration and feature comparison in a range of scales. To achieve this goal, we propose a \emph{non-skip U-Net} (nsUNet) that constructs a feature pyramid upon the U-shape architecture~\cite{ronneberger2015u}. In practice, nsUNet effectively avoids the production of shortcut solutions when performing the context restoration task. On the basis of nsUNet, we conduct pixel-level context restoration and siamese feature comparison in each level (i.e., scale) of the feature pyramid. In this way, the proposed framework helps improve the ability of self-supervised representations to recognize objects (e.g., lesions and organs in medical images) at different sizes and scales.

We summarize the contributions of this paper as follows:
\begin{itemize}
    \item We present an information preservation framework for advancing SSL in medical image analysis. In this framework, we unify the preservation of visual information in latent representations from three aspects: pixels, semantics, and scales. Towards this end, pixel restoration and feature comparison are conducted at different feature scales.
    \item We introduce non-skip U-Net (nsUNet) to construct the feature pyramid. Compared to the typical U-shape models in medical imaging~\cite{ronneberger2015u,cciccek20163d}, nsUNet maintains more feature scales and eliminates the usage of the widely adopted skip connections to avoid shortcut solutions to pixel restoration.
    \item Inspired by multi-crop~\cite{caron2020unsupervised}, we propose sub-crop to compare global volumes against local volumes. In order to mitigate the problem of the reduced mutual information between global and local views in 3D space, sub-crop restricts the cropping of local views within the 3D minimum bounding box of global views. Experiments on 3D medical images found that sub-crop is more effective than multi-crop in various downstream tasks.
    \item We conduct extensive and comprehensive experiments to validate the effectiveness of the proposed framework. We show that the unification of pixels, semantics, and scales can provide impressive performance under the pre-training/fine-tuning protocol. Specifically, the proposed framework outperforms both self-supervised and supervised counterparts in chest pathology classification, pulmonary nodule detection, abdominal organ segmentation, and brain tumor segmentation by substantial margins.
\end{itemize}
\begin{figure*}[t]
    \centering
    \subfloat[][Multi-scale pixel restoration]{\includegraphics[width=0.48\textwidth]{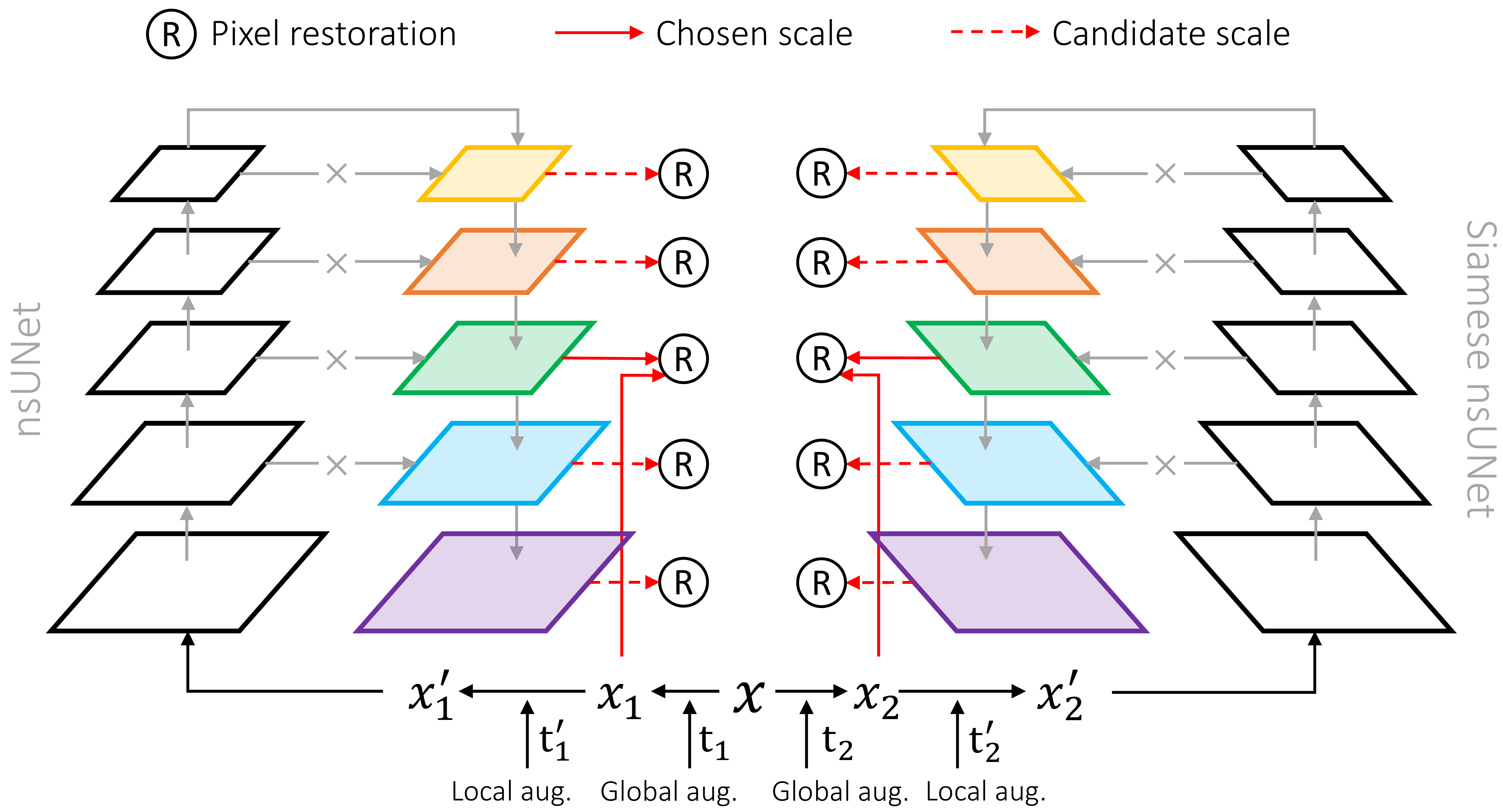}\label{overview_a}}\quad\quad
    \subfloat[][Multi-scale feature comparison]{\includegraphics[width=0.48\textwidth]{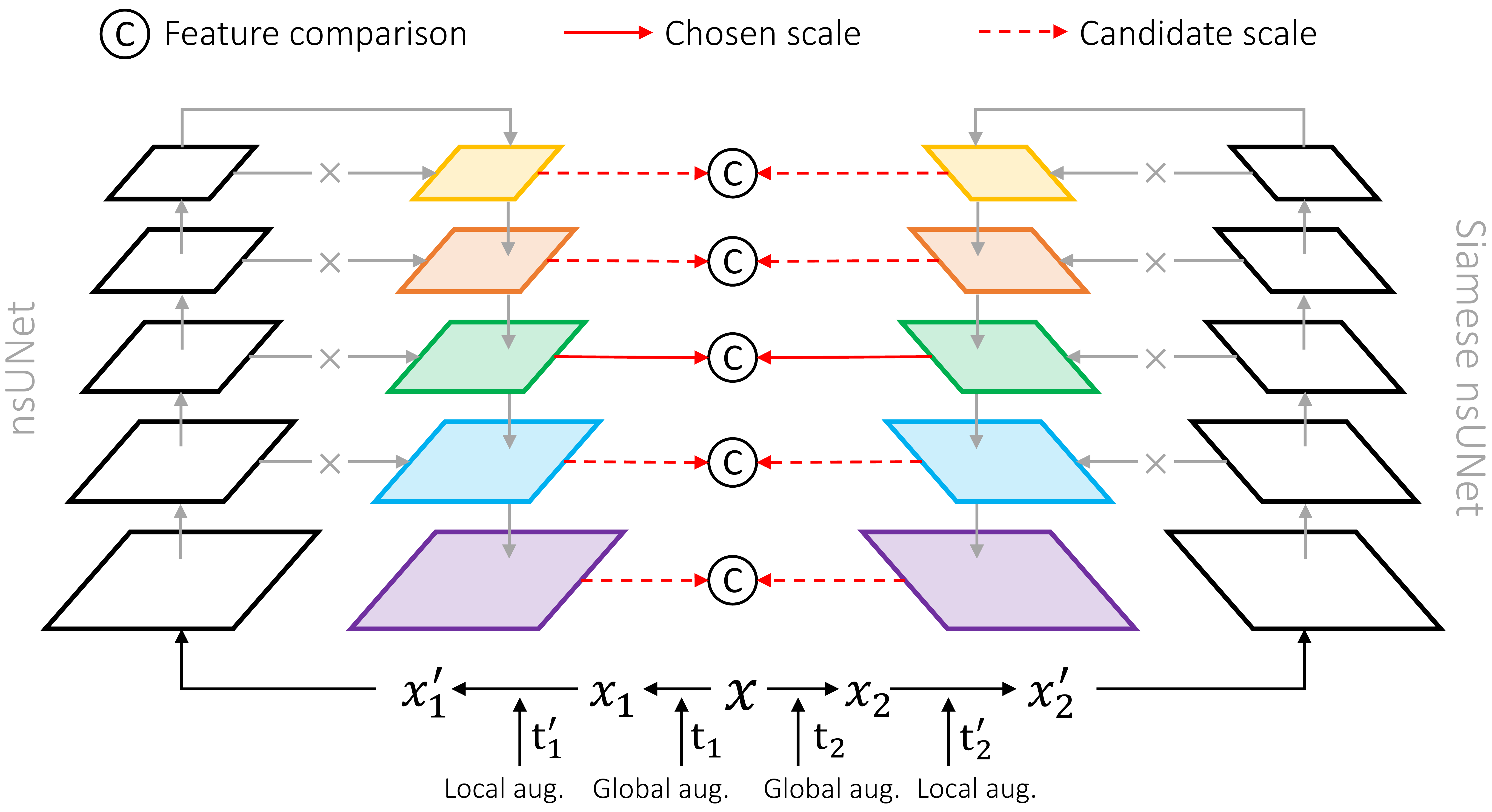}\label{overview_b}}    
    \caption{The overall structure of PCRLv2. PCRLv2 performs self-supervised visual learning on siamese feature pyramids. To achieve this goal, we propose non-skip U-Net (nsUNet). nsUNet consists of five feature scales and removes the skip connections to prevent network optimizers from finding shortcut solutions to context restoration. On the basis of nsUNet, we propose to decouple the preservation of pixel-level, semantic, and scale information into two tasks: (a) multi-scale pixel restoration; (b) multi-scale feature comparison. The rationale behind is to incorporate pixel details and semantics into features at different scales. During the training stage, we randomly choose a feature scale from the feature pyramid, on top of which we conduct pixel restoration and feature comparison. $x$ denotes a batch of input images. $t_1$ and $t_2$ stand for two distinct global augmentations, while $t_1^{\prime}$ and $t_2^{\prime}$ denote the successive local augmentations.}
    \label{overview}
\end{figure*}
The conference version of this paper (PCRLv1) was presented in \cite{zhou2021preservational}, which demonstrates the benefits of incorporating more pixel-level information besides the invariant and discriminative semantics obtained by contrastive learning. In this paper, we made significant and substantial modifications to PCRLv1, and we name the improved framework as \emph{PCRLv2} (i.e., \textsc{P}reservational \textsc{C}omparative \textsc{R}epresentation \textsc{L}earning). The modifications and improvements in PCRLv2 include but are not limited to \textbf{(i)} Besides local pixel-level and global semantic information, scale information is also preserved in self-supervised visual representations. The motivation behind is that although multiple feature scales have been considered in various vision tasks, they have not drawn much attention in SSL. PCRLv2 shows that introducing multi-scale latent representations can boost the transfer learning performance of SSL in downstream tasks. \textbf{(ii)} PCRLv2 simplifies the attentional pixel restoration and hybrid feature contrast operations of PCRLv1 into a concise multi-task optimization problem. As a result, PCRLv2 is simpler and easier to implement while achieving better performance, thus more practical. \textbf{(iii)} Compared to PCRLv1 that relies on the plain U-Net architecture~\cite{ronneberger2015u}, PCRLv2 conducts SSL on top of a new backbone, i.e., non-skip U-Net (nsUNet). There are two inherent advantages of nsUNet. First, the feature pyramid of nsUNet allows performing multi-scale pixel-level context restoration and semantic feature comparison. As a result, the unification of pixels, semantics, and scales produces more transferable visual representations. Second, nsUNet can effectively avoid the production of shortcut solutions, providing obvious performance gains over the use of the typical skip connections. \textbf{(iv)} We integrate the idea of multi-crop~\cite{caron2020unsupervised} in PCRLv2. Moreover, in 3D medical imaging, we propose sub-crop to produce reliable local views with increased mutual information by randomly cropping multiple local volumes within the 3D minimum bounding box of global views. In practice, we found that the proposed sub-crop has better pre-training performance than multi-crop. \textbf{(v)} In 5 classification/segmentation tasks, PCRLv2 provides more transferable pre-trained visual representations, not only surpassing previous self-supervised and supervised counterparts by substantial margins but also obviously outperforming PCRLv1 in all experiments.

\section{Related Work}
This section reviews related work in comparative SSL, including contrastive and non-contrastive methods, and lists SSL approaches that use context restoration as the pretext task. In the third part, we collect papers that emphasize the incorporation of multi-scale features in SSL.\\

\noindent\textbf{Comparative SSL methodologies.} One of the core ideas behind comparative SSL is to extract and encode invariant and discriminative semantics into representations via feature-level comparison. Hjelm~\etal~\cite{hjelm2018learning} proposed Deep InfoMax to maximize the mutual information between global and local feature vectors of the same input image using InfoNCE~\cite{oord2018representation}. Bachman~\etal~\cite{bachman2019learning} augmented InfoMax by conducting a global-local comparison on feature vectors of independently-augmented versions of each input. Tian~\etal~\cite{tian2020contrastive} increased the number of augmented views of each input and extended InfoNCE to multiple views. He~\etal~\cite{he2020momentum} presented Momentum Contrast (MoCo), which comprises a momentum encoder to maintain the consistency among positive and negative feature vectors. Different from~\cite{hjelm2018learning,bachman2019learning}, MoCo performs InfoNCE on top of global feature vectors only. Compared to MoCo, SimCLR removes the momentum architecture and defines InfoNCE on the output of a MLP with one hidden layer. Inspired by SimCLR, Chen~\etal~\cite{chen2020improved} proposed MoCov2, which improves MoCo with an additional MLP head and more augmentations. SwAV~\cite{caron2020unsupervised} replaces the feature vectors in InfoNCE with cluster assignments and introduces the multi-crop strategy to increase the number of views of an image with affordable computational overhead. Grill~\etal~\cite{grill2020bootstrap} proposed BYOL (bootstrap your own latent), which eliminates the use of InfoNCE in SSL by distilling semantics from positive pairs only. Based on BYOL, Chen~\etal~\cite{chen2021exploring} further removed the restriction of the momentum architecture and introduced a simple siamese learning framework named SimSiam. In practice, SimSiam produces comparable results to MoCov2 in various downstream tasks. Recently, Zbontar~\etal~\cite{zbontar2021barlow} simplified SimSiam by measuring the cross-correlation matrix between the siamese global feature vectors and trying to make this matrix close to the identity.

Comparative SSL, especially InfoNCE-based methodology, has also been widely adopted in medical image analysis. Zhou~\etal~\cite{zhou2020comparing} proposed to integrate mixup~\cite{zhang2017mixup} into MoCov2, increasing the diversity of both positive and negative samples in InfoNCE. Taleb~\etal~\cite{taleb20203d} developed 3D versions of existing SSL techniques and compared 2D and 3D SSL approaches on downstream tasks. Azizi~\etal~\cite{azizi2021big} incorporated multi-instance learning into SimCLR, which helps utilize multiple views of each patient. Around the same time, Vu~\etal~\cite{vu2021medaug} developed a method to select positive pairs coming from views of the same patient and used this strategy to improve MoCov2. There are also a number of approaches~\cite{chaitanya2020contrastive,you2021momentum,you2022simcvd} that tailored comparative SSL for semi-supervised medical image segmentation. 

However, the methodologies mentioned above fail to address the importance of integrating pixel-level information into the high-level representations with rich semantics, which is the primary focus of the proposed PCRL.\\

\noindent\textbf{Context restoration for preserving pixel-level information.} Restoring original context has been treated as an important pretext task in SSL. Pathak~\etal~\cite{pathak2016context} first time conducted self-supervised feature learning by recovering masked input images. Larsson~\etal~\cite{larsson2016learning} and Zhang~\etal~\cite{zhang2017split} performed SSL on pixels via predicting RGB color values. For medical images, Chen~\etal~\cite{chen2019self} extended the approach in~\cite{pathak2016context} with swapped image patches. Zhou~\etal~\cite{zhou2021models} showed that adding more augmentations to input images brings benefits to SSL. Tao~\etal~\cite{tao2020revisiting} presented a volume-wise context transformation for 3D medical images. Different from the approaches mentioned above, Henaff~\cite{henaff2020data} proposed to predict the next context feature vectors following an auto-regressive manner. 

We can see that context restoration is more prevalent in medical imaging than in natural images from the above. The underlying reason is that medical imaging tasks require more pixel-level information to make fine-grained yet accurate decisions. On the other hand, we observe that comparative SSL can produce representations with richer semantics. Thus, it can be beneficial to build a SSL framework that simultaneously integrates pixel-level and semantic information. As far as we are concerned, none of these context restoration based approaches incorporate such a combination.\\

\noindent\textbf{Multi-scale features in SSL.} Although multi-scale features have not drawn much attention in existing SSL research, it has already been treated as an implicit yet effective regularization method for SSL in some methodologies. Deep InfoMax~\cite{hjelm2018learning} contrasts high-level feature vectors with low-level feature maps using InfoNCE. To improve Deep InfoMax, Bachman~\etal~\cite{bachman2019learning} proposed to contrast global and local feature vectors on multiple levels. In medical image analysis, preserving scale information becomes essential, as pathologies may show different characteristics on different scales. In~\cite{chaitanya2020contrastive}, a local contrastive loss is introduced to learn distinctive representations of local regions that are helpful to per-pixel segmentation. At the same time, global feature vectors are used to distill discriminative semantics for classification tasks. A similar idea has also been used in image registration~\cite{liu2021same} and one-shot segmentation~\cite{zhou2021generalized}, where global and local feature vectors are employed to provide information on semantics and position, respectively. 

However, most of these methods only perform SSL on two scales, i.e., one global and one local, which cannot fully capture multi-scale information. Besides, although these approaches emphasize the benefit of introducing local information to SSL, they do not exploit pixel-level information that is helpful to encode locality. In contrast, this paper proposes a unified framework that can simultaneously preserve semantic, pixel-level, and scale information. 

\section{Methodology}
We provide an overview of PCRLv2 in Fig.~\ref{overview}. Suppose $x$ denotes a batch of input images. We introduce cascaded augmentations to distort $x$ in global and local views, respectively. To be specific, the first-stage augmentations ($t_1$ and $t_2$ in Fig.~\ref{overview}) mainly consist of global transformations, such as flip and rotation, whose goal is to distort the semantics of input images from a global perspective. In comparison, the second-stage augmentations ($t_1^{\prime}$ and $t_2^{\prime}$ in Fig.~\ref{overview}) comprise local pixel-level transformations, such as random noise and gaussian blur, which are leveraged to perturb the local semantics. After two-stage augmentations, the finally augmented images $x_1^{\prime}$ and $x_2^{\prime}$ are passed to siamese networks to perform pixel restoration and feature comparison, while the results of applying $t_1$ and $t_2$ to $x$, i.e., $x_1$ and $x_2$, serve as the ground truth targets for the pixel restoration task (as shown in Fig.~\ref{overview_a}). 

We perform SSL on the feature pyramid to encode multi-scale visual representations. Following the standard practice in medical image processing, we build feature pyramids using a U-shape model named non-skip U-Net (nsUNet). Compared to the typical U-Net architecture~\cite{ronneberger2015u,cciccek20163d}, nsUNet has more feature scales and completely removes skip connections, both of which we empirically found helpful in producing better pre-trained representations. During the training stage, one scale is first randomly chosen from all five feature scales, after which we conduct pixel restoration and feature comparison on the siamese feature maps at the chosen scale. After the pre-training stage, we fine-tune the encoder of nsUNet on various downstream tasks.

\subsection{Feature pyramid in non-skip U-Net}
U-Net and its series~\cite{ronneberger2015u,cciccek20163d,isensee2021nnu} have been known in medical imaging for their abilities to handle image segmentation tasks. The most distinctive characteristic of these models is the skip connection that connects equal-resolution low- and high-level feature maps. The critical insight is to recover the spatial information lost in down-sampling operations of the encoder network, such as strided pooling or convolution. U-shape models use a feature pyramid to progressively incorporate multi-scale details brought by skip connections into high-level semantics, making the U-shape architecture an ideal choice for conducting context restoration.

In this paper, we explore the potential of U-shape architecture in SSL from two perspectives: deeply fusing semantic and pixel-level information by removing the skip connections and introducing multi-scale latent representations by conducting SSL on the feature pyramid. For the first perspective, we empirically found that skip connections provide shortcuts for context restoration, as the low-level feature maps contain rich, high-resolution pixel-level details. This characteristic does contribute to the restoration of context. However, it may prevent the high-level latent representations (with rich semantics) from incorporating more pixel-level information because the task of providing pixel-level details is assigned to low-level feature maps. To address this point, we remove the skip connections in U-shape architecture and propose non-skip U-Net (nsUNet). nsUNet relies on high-level representations without any skip connections to restore pixel-level details. In this way, the semantic and pixel-level information can be deeply fused. Meanwhile, the inherent multi-scale feature maps of nsUNet offer the opportunity to construct a feature pyramid, on top of which SSL can be conducted in multiple scales simultaneously.

\begin{figure}[t]
    \centering
    \includegraphics[width=0.85\columnwidth]{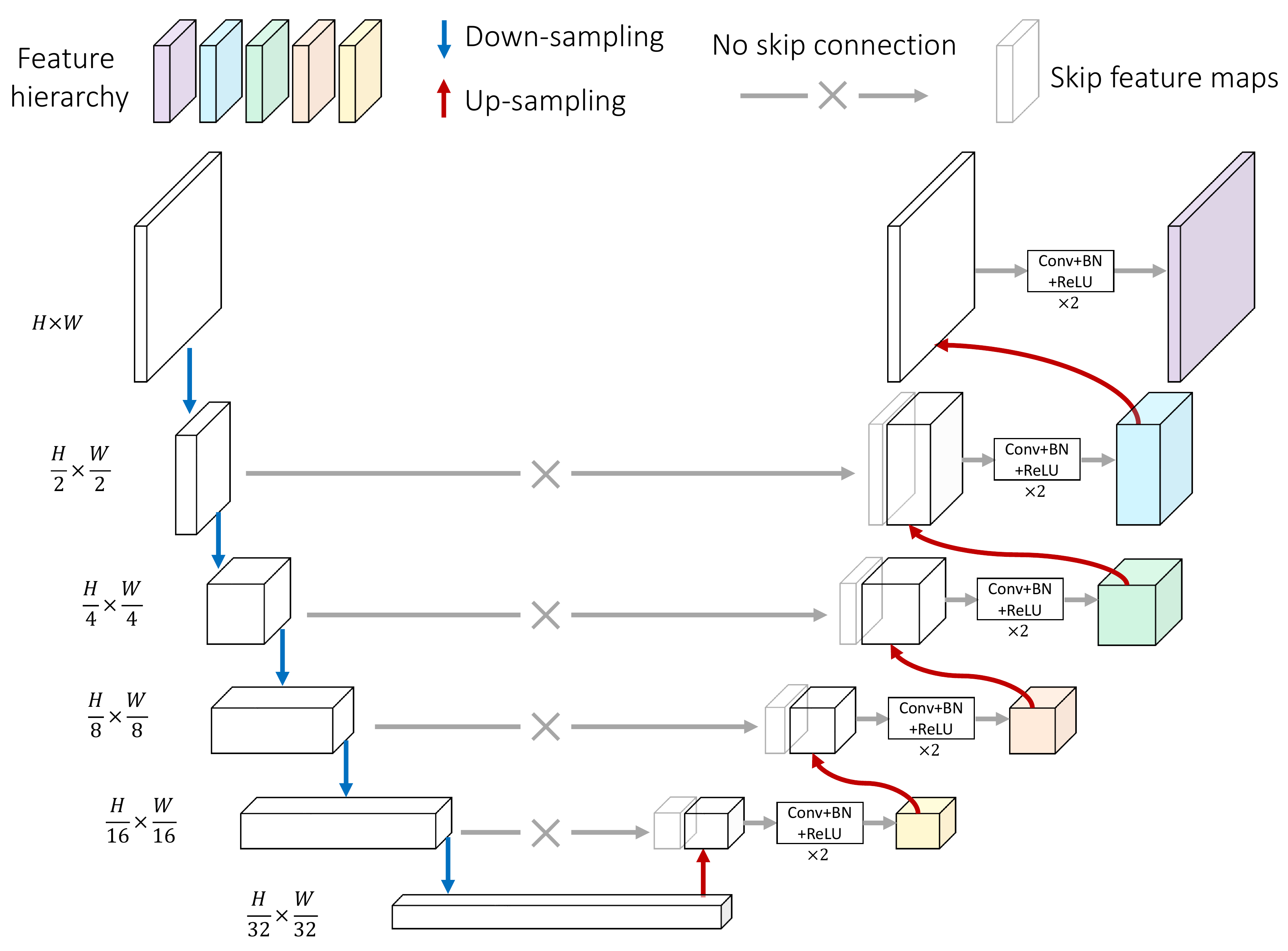}
    \caption{The architecture of non-skip U-Net (nsUNet). In comparison to previous U-Net series, nsUNet removes skip connections, and the associated skip feature maps to prevent shortcut solutions to the pixel restoration and feature comparison tasks. Besides, nsUNet consists of five levels of feature maps (denoted with different colors), where two self-supervised tasks are further conducted. Note that this is a 2D illustration of nsUNet.}
    \label{nsUNet}
\end{figure}

Fig.~\ref{nsUNet} presents the architecture of nsUNet. The feature pyramid in nsUNet comprises five levels, ranging from low resolution (the down-sampling rate is 32) to full resolution (no down-sampling). For 2D input data, we use ResNet-18~\cite{he2016deep} as the encoder, while for 3D input volumes, we build the encoder following~\cite{cciccek20163d}. As illustrated in Fig.~\ref{nsUNet}, the decoder of nsUNet maintains a shared architecture across all pyramid levels, which can be summarized as:
\begin{align}
    \begin{split}
        \mathcal{F}_i = \text{Conv-BN-ReLU}(\text{Conv-BN-ReLU}(\text{Up}(\mathcal{F}_{i-1})),
    \end{split}
\end{align}
where $i\in \{1,2,3,4,5\}$. $\mathcal{F}_0$ denotes the output of the bottleneck block, which has the lowest spatial resolution (down-sampling rate=32). \emph{Up} represents the up-sampling operation. \emph{Conv-BN-ReLU} stands for a sequence of operations, including convolution (kernel size=3), batch normalization (BN), and ReLU activation. As a result, the bag of feature maps $\{\mathcal{F}_1,\mathcal{F}_2,\mathcal{F}_3,\mathcal{F}_4,\mathcal{F}_5\}$ is then forwarded to following task-dependent heads to perform pixel restoration and feature comparison, respectively and simultaneously.

\subsection{Multi-scale pixel restoration}
\begin{figure*}[t]
    \centering
    \subfloat[][Architectural details of the pixel restoration head]{\includegraphics[height=0.43\columnwidth]{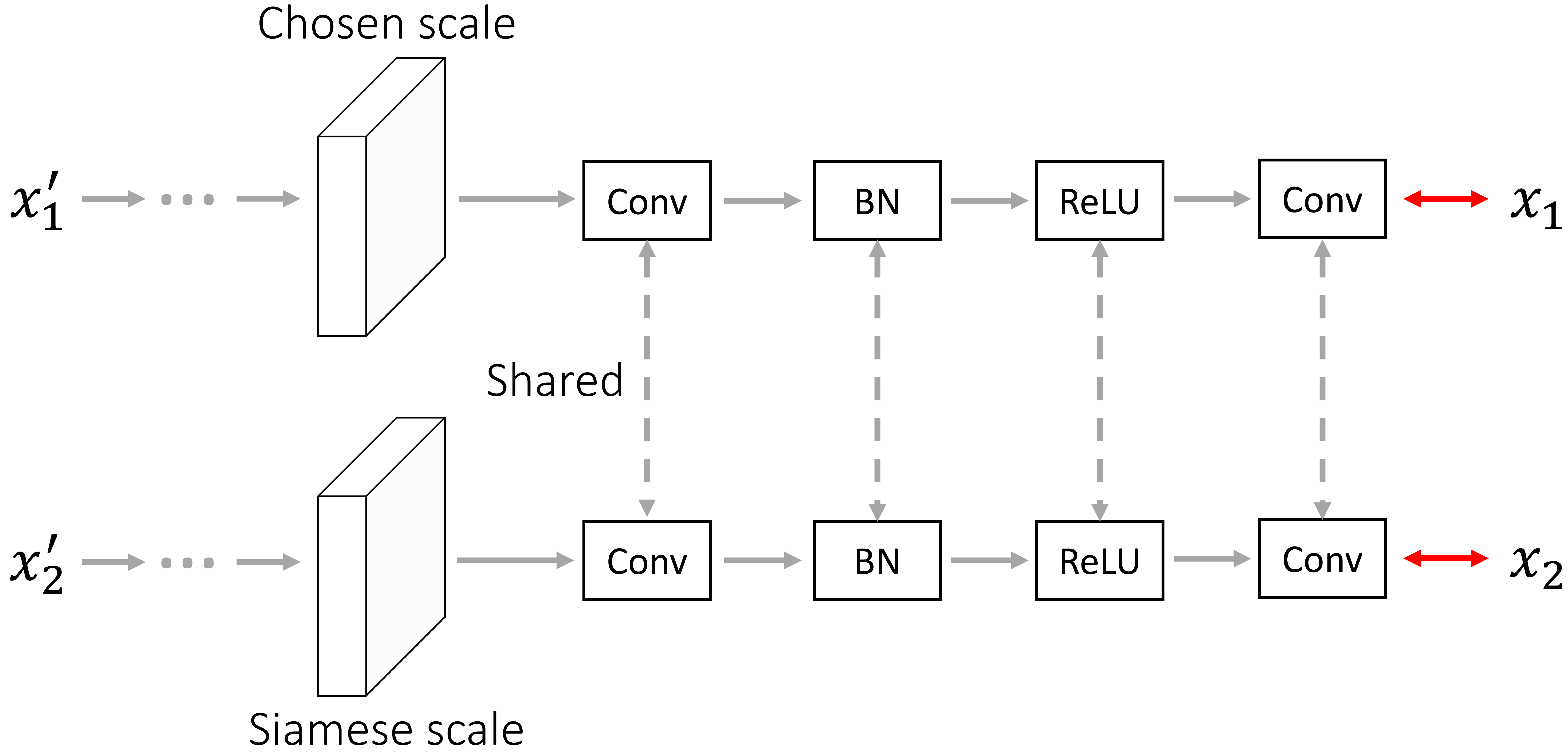}\label{task1}}\quad
    \subfloat[][Architectural details of the feature comparison head]{\includegraphics[height=0.43\columnwidth]{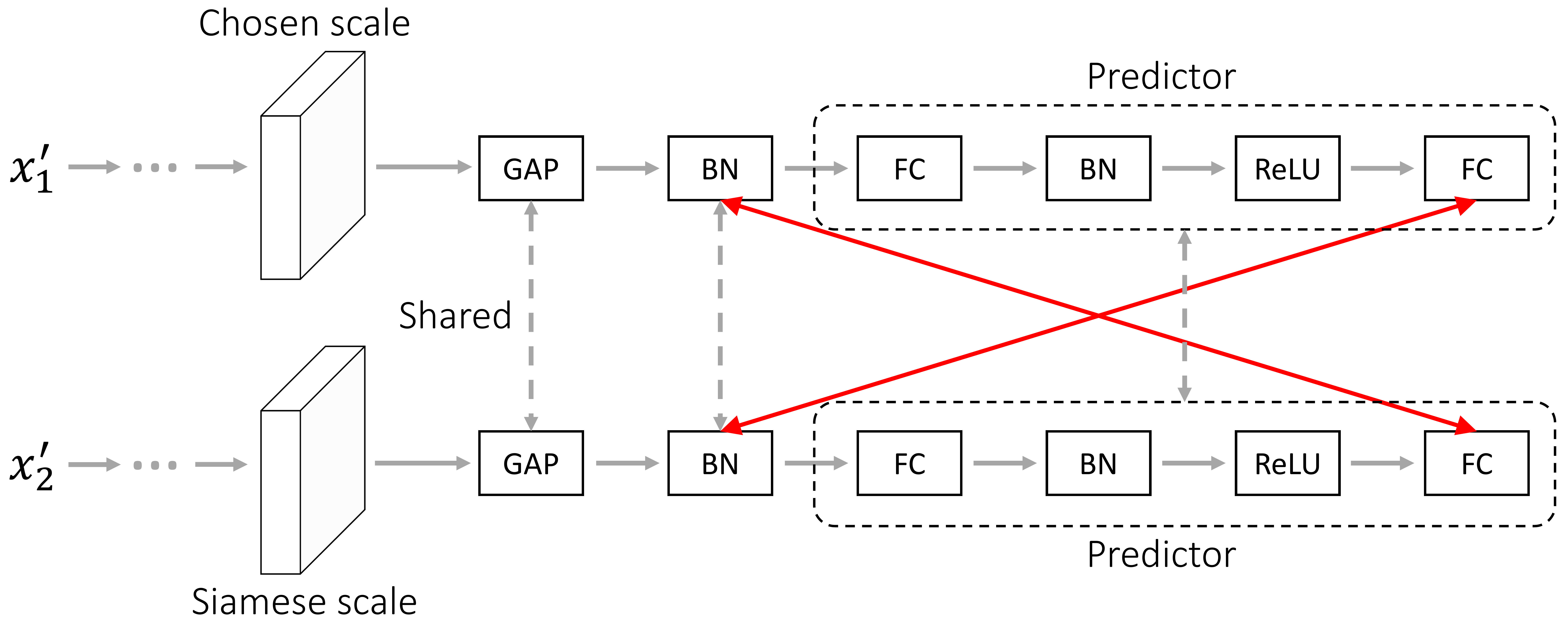}\label{task2}} 
    \caption{Architectural details of the pixel restoration and feature comparison heads. \textbf{Conv}, \textbf{BN}, \textbf{GAP}, and \textbf{FC} denote the convolution, batch normalization, global average pooling, and fully-connected layers, respectively. The kernel size of all convolution layers is 3, and the convolution stride is set to 1. Note that each pair of siamese feature maps share one pixel restoration head and one feature comparison head, while different feature scales employ distinct task heads.}
    \label{tasks}
\end{figure*}
As the name implies, multi-scale pixel restoration aims to preserve pixel-level and scale information in latent visual representations simultaneously. To achieve this goal, we ask the network to recover the exact pixel-level details across different scales, where each pair of siamese feature maps share one pixel restoration head. In contrast, PCRLv1 only restores pixel details at the full resolution, which inevitably loses multi-scale properties in learned representations.

As shown in Fig.~\ref{task1}, the input images $x_1^{\prime}$ and $x_2^{\prime}$ are intentionally corrupted via various pixel-level augmentations, such as guassian blur and random noise. For each training iteration, we first randomly choose a feature scale $\mathcal{F}_i$ from $\{\mathcal{F}_1,\mathcal{F}_2,\mathcal{F}_3,\mathcal{F}_4,\mathcal{F}_5\}$. Then, we pass $\mathcal{F}_i$ to the pixel restoration head $f^\text{R}_i(\cdot)$ for the $i$-th scale, whose internal processing procedure can be summarized as:
\begin{align}
    \begin{split}
        f^\text{R}_i(\mathcal{F}_i) = \text{Conv}(\text{Conv-BN-ReLU}(\mathcal{F}_i)),
    \end{split}
\end{align}
where all convolution layers use a kernel size of 3 and a stride of 2. Similarly, we apply the shared pixel restoration head to the paired siamese feature map $\mathcal{F}_i^s$ to acquire the prediction output $f^R_i(\mathcal{F}_i^s)$:
\begin{align}
    \begin{split}
        f^\text{R}_i(\mathcal{F}^s_i) = \text{Conv}(\text{Conv-BN-ReLU}(\mathcal{F}^s_i)),
    \end{split}
\end{align}

Lastly, we employ the mean square error (MSE) loss to measure the reconstruction errors between $f^\text{R}_i(\mathcal{F}_i)$ and $x_1$. For the siamese feature pyramid, we apply MSE loss to $f^\text{R}_i(\mathcal{F}_i^s)$ and $x_2$. The cost function $\mathcal{L}^\text{R}$ of the pixel restoration task in each training iteration (with mini-batch optimization) is as follows:
\begin{align}
    \begin{split}
        \mathcal{L}^\text{R} = \sum_{\substack{j=1,\\\forall i\in H}}^N \mathbbm{1}_{[i\Equal\Equal j]}\ [\text{MSE}(f^\text{R}_i(\mathcal{F}_i),x_1)+\text{MSE}(f^\text{R}_i(\mathcal{F}_i^s), x_2)],
    \end{split}
    \label{mse_eq}
\end{align}
where $N=5$ denotes the number of scales in each feature pyramid. $H=\{1,2,3,4,5\}$ stands for the scale index. $\mathbbm{1}_{[i\Equal\Equal j]}$ is an indicator function, which is equal to 1 when $i$==$j$ is true (otherwise, 0). The explanation of $\mathcal{L}^\text{R}$ can be summarized as: \textbf{(i)} randomly choose a feature scale $\mathcal{F}_i$ from all five scales; \textbf{(ii)} pass $\mathcal{F}_i$ and its siamese feature map $\mathcal{F}_i^s$ to the shared task head $f^\text{R}_i(\cdot)$; \textbf{(iii)} calculate the MSE loss between the outputs of $f^\text{R}_i(\cdot)$ and uncorrupted images $\{x_1, x_2\}$. By reconstructing the same targets $x_1$/$x_2$ across different feature scales, $\mathcal{L}^\text{R}$ can encode the pixel-level information into multi-scale latent visual representations.

\subsection{Multi-scale feature comparison}
PCRLv1 employs a hybrid way to conduct contrastive learning with the help of the momentum encoder~\cite{he2020momentum} and mixup~\cite{zhang2017mixup}. However, this contrastive deployment is complex, making PCRLv1 heavy, thus troublesome to implement and improve. To address these issues, PCRLv2 replaces the hybrid contrastive strategies in PCRLv1 with the multi-scale comparison. Inspired by \cite{chen2021exploring}, multi-scale comparison conducts SSL with siamese learning, whose key operation is to attract the same image's siamese views. Different from \cite{chen2021exploring} that conducts feature comparison on one scale, we propose to preserve the discriminative semantics across different feature scales, which forces the model to preserve multi-scale self-supervised representations. In the following, we provide technical details of performing the multi-scale comparison.

Given the feature maps at a randomly chosen scale $\mathcal{F}_i$, we pass them through a global average pooling layer and a shared batch normalization layer (as shown in Fig.~\ref{task2}) to acquire 1D representations $\mathbf{v}_i$:
\begin{align}
    \mathbf{v}_i &= \text{BN}(\text{GAP}(\mathcal{F}_i)).
\end{align}
We can get $\mathbf{v}^s_i$ by processing the siamese feature maps $\mathcal{F}^s_i$ in a similar way.

Next, we forward $\mathbf{v}_i$ to the shared predictor $f_\text{P}(\cdot)$, whose architecture is displayed in Fig.~\ref{task2} and can be summarized as:
\begin{align}
    f_\text{P}(\mathbf{v}_i) &= \text{FC}(\text{FC-BN-ReLU}(\mathbf{v}_i)).
\end{align}
where \emph{FC} denotes the fully-connected layer. \emph{FC-BN-ReLU} stands for a sequence of layers, which are the fully-connected layer, batch normalization layer, and ReLU activation. Similarly, we can acquire $f_\text{P}(\mathbf{v}^s_i)$ by passing $\mathbf{v}^s_i$ to the same predictor.

We measure the similarity between siamese feature vectors with the cosine similarity:
\begin{align}
    \begin{split}
        \text{cos}(\mathbf{v}_i, f_\text{P}(\mathbf{v}^s_i))=\frac{\mathbf{v}_i}{\left\|\mathbf{v}_i\right\|_{2}} \cdot \frac{f_\text{P}(\mathbf{v}^s_i)}{\left\|f_\text{P}(\mathbf{v}^s_i)\right\|_{2}},
    \end{split}
\end{align}
where $||\cdot||_2$ denotes the L2 normalization. Symmetrically, we calculate $\text{cos}(f_\text{P}(\mathbf{v}_i), \mathbf{v}^s_i)$ as follows:
\begin{align}
    \begin{split}
        \text{cos}(f_\text{P}(\mathbf{v}_i), \mathbf{v}^s_i))=\frac{f_\text{P}(\mathbf{v}_i)}{\left\|f_\text{P}(\mathbf{v}_i)\right\|_{2}} \cdot \frac{\mathbf{v}^s_i}{\left\|\mathbf{v}^s_i\right\|_{2}}.
    \end{split}
\end{align}
Finally, the cost function $\mathcal{L}_\text{C}$ of multi-scale feature comparison can be summarized as:
\begin{align}
    \begin{split}
        \mathcal{L}^\text{C} = \sum_{\substack{j=1,\\\forall i\in H}}^N -\frac{1}{2}\mathbbm{1}_{[i\Equal\Equal j]}\ [\text{cos}(\textit{sg}(&\mathbf{v}_i), f_\text{P}(\mathbf{v}^s_i))\\ &+\text{cos}(f_\text{P}(\mathbf{v}_i), \textit{sg}(\mathbf{v}^s_i))].
    \end{split}
    \label{loss2}
\end{align}
$N=5$ denotes the number of feature scales. $H=\{1,2,3,4,5\}$ stands for the scale index. Following \cite{chen2021exploring}, we apply the stop-gradient operation (denoted as $\textit{sg}$) in Eq.~\ref{loss2} to prevent the network optimizer from finding shortcut solutions.

Minimizing $\mathcal{L}^\text{C}$ requires the model to maximize the similarity between siamese latent features across all feature scales. In this way, scale invariance can be implicitly incorporated into the preserved latent semantics.

\subsection{From multi-crop to sub-crop}
\begin{figure}[t]
    \centering
    \includegraphics[width=1.0\columnwidth]{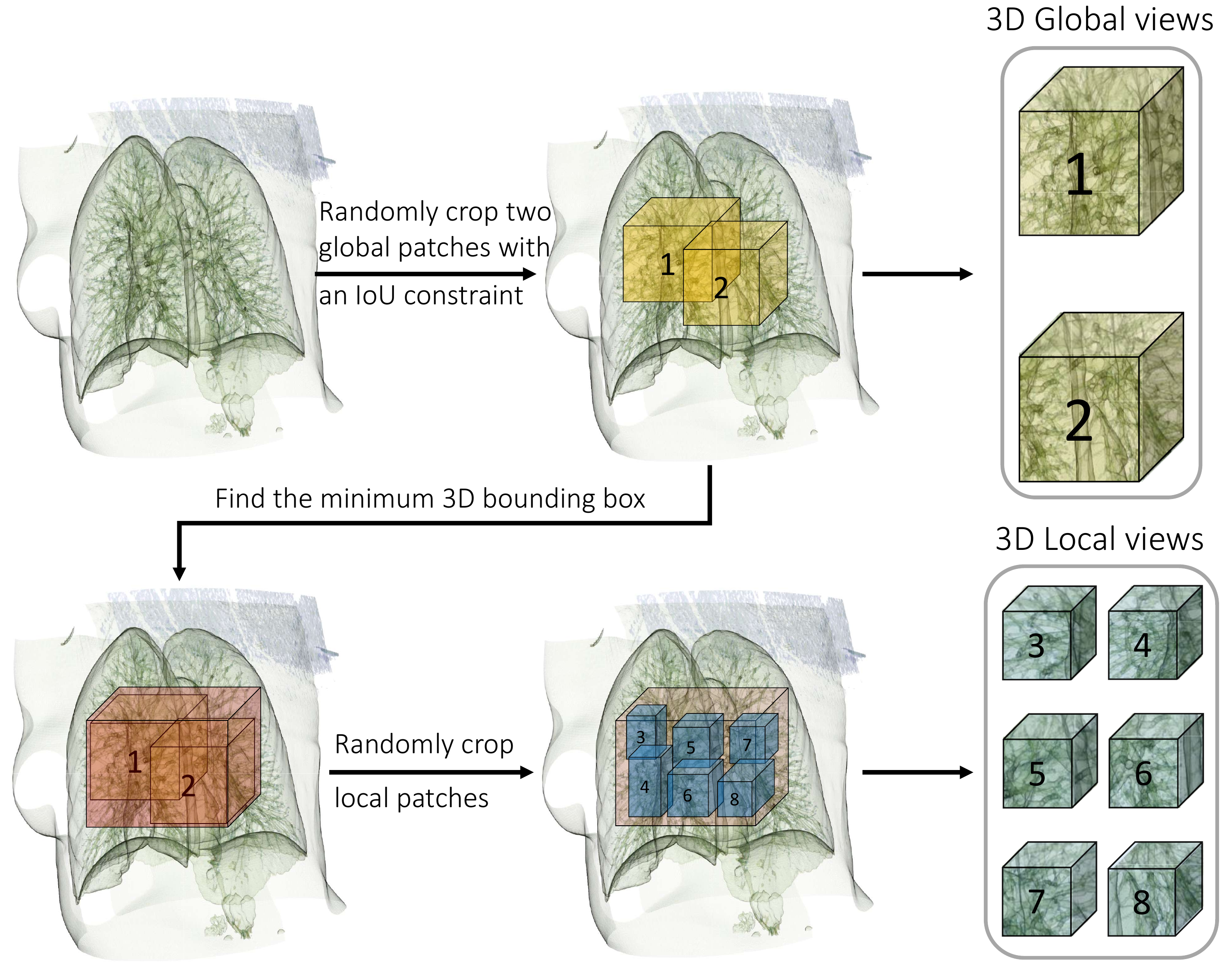}
    \caption{Illustration of sub-crop. Given a 3D local volume, we first randomly crop two large patches, where an intersection over union (IoU) constraint is applied to guarantee that two patches are partly overlapped. These two large patches are considered as $x_1$ and $x_2$ in Fig.~\ref{overview} and will be passed to the siamese architecture to conduct the following multi-scale pixel restoration and feature comparison tasks. To acquire local views, we compute the minimum 3D bounding box of two large patches, after which random crop is applied to extract multiple local patches. Finally, we reshape these local patches to a fixed size and forward them to the network to extract local representations.}
    \label{sub_crop}
\end{figure}
Multi-crop~\cite{caron2020unsupervised} has been known as a helpful strategy to improve SSL performance in natural images, which increases the number of input views by sampling several standard resolution crops and more low-resolution crops from the original input. One key insight behind multi-crop is to capture relations between parts of a scene or an object, while low-resolution views ensure a controllable increase in the computational cost. 

When applied to medical images, multi-crop works well in 2D X-ray data but leads to the non-convergence of the model in 3D volume data (such as CT and MRI). After careful investigation, we found the root of this problem lies in the contradiction between the limited input size and many candidate crops in three-dimensional space. Specifically, on the one hand, we cannot afford large-sized 3D inputs because processing them with 3D deep models often costs dramatic GPU memory. On the other hand, if we overly reduce the size of 3D inputs, the sampled views would be too dispersed to guarantee the model capture the local-global associations. 

To mitigate the above issue, we introduce \emph{sub-crop} to replace multi-crop in 3D medical images. The core idea of \emph{sub-crop} is straightforward: reducing the sampling space. As illustrated in Fig.~\ref{sub_crop}, sub-crop mainly consists of three steps: (i) randomly crop two extensive global views with an IoU constraint; (ii) find the minimum 3D bounding box over the cropped global patches; (iii) randomly crop multiple local patches within the 3D bounding box. There are two critical operations in sub-crop: the constraint of IoU on global views and the sampling of local patches within the minimum bounding box. In practice, the first operation guarantees the global-global association by ensuring the overlap between large patches larger than a fixed threshold. The second operation mitigates the disperse problem of local views and helps the model to discover local-global relations.

\subsection{Overall training objective}
After applying multi-crop/sub-crop to medical images, we can acquire two global views $\{\mathbf{g}_1, \mathbf{g}_2\}$ and $\hat{N}$ local views $\{\mathbf{l}_1,\mathbf{l}_2,...,\mathbf{l}_{\hat{N}}\}$. For clarification, we denote the associated inputs in notations of loss functions. For instance, $\mathcal{L}^{C}(\mathbf{g}_1, \mathbf{g}_2)$ means we calculate $\mathcal{L}^{C}$ on top of the extracted siamese representations of two global views, where $\mathbf{g}_1$ and $\mathbf{g}_2$ can be regarded as a pair of siamese images. At last, the overall training objective of PCRLv2 can be formalized as follows:
\begin{align}
    \begin{split}
        \mathcal{L}^{\text{Total}}(\mathbf{g}_1,\mathbf{g}_2,\mathbf{l}_1,...,\mathbf{l}_{\hat{N}})=&\mathcal{L}^\text{R}(\mathbf{g}_1, \mathbf{g}_2)  + \mathcal{L}^{\text{C}}(\mathbf{g}_1, \mathbf{g}_2) \\ 
         &+ \sum_{m\in \{1,2\}} \sum_{k=1}^{\hat{N}} \mathcal{L}^{\text{C}}(\mathbf{l}_k, \mathbf{g}_m).
    \end{split}
\end{align}
There are three terms in $\mathcal{L}^{\text{Total}}$: $\mathcal{L}^\text{R}(\mathbf{g}_1, \mathbf{g}_2)$, $\mathcal{L}^{\text{C}}(\mathbf{g}_1, \mathbf{g}_2)$, and $\sum_{m\in \{1,2\}} \sum_{k=1}^{\hat{N}} \mathcal{L}^{\text{C}}(\mathbf{l}_k, \mathbf{g}_m)$. The first term is designed to preserve pixel-level details in multi-scale learned representations. The second term addresses the importance of encoding multi-scale semantics into latent features. The last term aims to capture the multi-scale global-local semantic relations.

\subsection{Short discussion: PCRLv2 vs. PCRLv1}
\noindent\textbf{Simpler.} PCRLv1 combines the context restoration and comparative SSL via transformation-conditioned attention and cross-model mixup. These two components make the framework heavy, less intuitive, and not easy to implement. Compared to PCRLv1, PCRLv2 exploits a simpler yet more intuitive design to incorporate pixel-level and semantic information via multi-scale learning. As aforementioned, PCRLv2 can be formulated as a simple multi-task optimization problem whose objective function maximizes the preservation of multi-level information in latent visual representations. These characteristics make it easier for both implementation and potential expansion.\\

\noindent\textbf{Faster.} PCRLv1 makes heavy use of mixup (to both inputs and features) in its implementation, which is found to deliver performance gains. In PCRLv2, we eliminate mixup strategies and cut the training time in half. In addition, PCRLv2 requires less running memory in GPUs during the training stage, making it more practical in real-world scenarios.

\section{Experiments}
In this section, we first conduct thorough ablation studies to investigate the influence of different modules in PCRLv2. Then, we evaluate the effectiveness of PCRLv2 on both 2D and 3D medical imaging tasks, including chest pathology classification, pulmonary nodule detection, abdominal organ segmentation, and brain tumor segmentation. For model evaluation, we follow the \emph{pre-training (on source data)}$\rightarrow$\emph{fine-tuning (on target data)} protocol and employ two settings, which are semi-supervised learning and transfer learning. In the first setting, the source and target data come from the same dataset. Specifically, we first pre-train the model using all training data without labels, and then fine-tune the pre-trained model with limited annotations. As for transfer learning (the second setting), we pre-train and fine-tune the model on different datasets. Different from semi-supervised learning, we fine-tune the pre-trained model with both limited and full annotations in transfer learning.

\subsection{Datasets}
\noindent\textbf{NIH ChestX-ray (2D)}~\cite{wang2017chestx} is made up of 112,120 X-ray scans from 30,805 patients. There are fourteen different chest pathologies in NIH ChestX-ray, including atelectasis, cardiomegaly, consolidation, edema, effusion, emphysema, fibrosis, hernia, infiltration, mass, nodule, pleural thickening, pneumonia, and pneumothorax. The labels of radiographs were automatically extracted from associated radiology reports using natural language process (NLP) techniques. We use NIH ChestX-ray in semi-supervised learning in our experiments and treat it as the target dataset in transfer learning. \\

\noindent\textbf{CheXpert (2D)}~\cite{irvin2019chexpert} involves 224,316 chest radiographs from 65,240 patients for the presence of 14 common chest radiographic observations: no finding, enlarged cardio, cardiomegaly, lung opacity, lung Lesion, edema, consolidation, pneumonia, atelectasis, pneumothorax, pleural effusion, pleural other, fracture, and support devices. Similar to NIH ChestX-ray, an NLP labeler was developed to detect the presence of 14 observations in radiology reports automatically. In practice, CheXpert serves as the source data in transfer learning.\\

\noindent\textbf{LUNA (3D)}~\cite{setio2017validation} was collected for the automatic detection of pulmonary nodules, which involves 888 annotated thoracic computed tomography (CT) scans. LUNA is a cherry-picked subset of LIDC-IDRI~\cite{armato2011lung}, which excludes scans with a slice thickness greater than 3mm, inconsistent slice spacing, or missing slices. In the 888 scans, a total of 5,855 annotations were made by the radiologists, where only nodules $\geq$ 3mm are categorized as relevant lesions, and at least one radiologist checks each nodule. On LUNA, we perform semi-supervised learning and transfer learning experiments. For transfer learning, LUNA is mainly used for self-supervised pre-training. \\

\noindent\textbf{LiTS (3D)}~\cite{bilic2019liver} releases 131 abdominal CT Volumes and associated annotations for training and validation. There are two types of labels in LiTS: the liver and tumor. In this paper, we only utilize the ground truth masks of the liver to evaluate the effectiveness of various SSL algorithms. The task on LiTS is abdominal organ segmentation, where LiTS is used for fine-tuning in transfer learning.\\

\noindent \textbf{BraTS (3D)} has been known as a series of challenges in brain tumor segmentation. In this paper, we perform experiments on the released 351 magnetic resonance imaging (MRI) scans of BraTS 2018. There are three classes in BraTS: whole tumor (WT), tumor core (TC), and enhancing tumor (ET). Similar to the role of LiTS, BraTS serves as the target data in transfer learning.

\subsection{Baselines}
A variety of SSL baselines are included in our extensive experiments, which can be roughly divided into three categories: 2D specific methods, 3D specific approaches, and generic (2D \& 3D) methodologies. Details of baselines in each category are listed below.\\

\noindent\textbf{2D specific SSL methodologies} consist of ImageNet-based pre-training (IN)~\cite{deng2009imagenet}, Comparing to Learn (C2L)~\cite{zhou2020comparing}, and Simple Siamese Learning (SimSiam)~\cite{chen2021exploring}. IN is the most widely adopted pre-training methodology, which conducts supervised pre-training on one of the biggest natural image datasets, i.e., ImageNet~\cite{deng2009imagenet}. C2L is a recently proposed SSL approach based on momentum contrast (i.e., MoCov1~\cite{he2020momentum} and MoCov2~\cite{chen2020improved}). SimSiam is a simple siamese SSL framework that eliminates the barrier of negative samples in contrastive learning and the use of a momentum encoder in BYOL~\cite{grill2020bootstrap}. Besides, we compare PCLRv2 against SimSiam to highlight the significance of the preserved pixel-level information and multi-scale features.\\

\noindent\textbf{3D specific SSL methodologies} include Rubik's cube++~\cite{tao2020revisiting} and 3D-CPC~\cite{taleb20203d}. Rubik's cube++ is the most recent SSL approach built on top of context restoration for 3D medical images. It adopts a volume-wise transformation for context permutation. In comparison, 3D-CPC is based on contrastive predictive encoding~\cite{henaff2020data}, a variation of contrastive learning, and demonstrates the most superior performance among different SSL approaches investigated in~\cite{taleb20203d}.\\

\noindent\textbf{Generic SSL methodologies} involve train from scratch (TS), Model Genesis (MG)~\cite{zhou2021models}, TransVW~\cite{haghighi2021transferable}, and PCRLv1~\cite{zhou2021preservational} (the conference version of our approach). MG resorts to aggressive augmentations to generate corrupted input images, based on which the model is asked to restore the original inputs. TransVW improves MG by appending an intermediate classification head to encode anatomical patterns explicitly. PCRLv1 first proposes simultaneously preserving semantic and pixel-level information in SSL.
\subsection{Implementation details}
\noindent\textbf{Dataset pre-processing for pre-training.} On NIH ChestX-ray and CheXpert, each input image is resized to 224$\times$224 after random crop. On LUNA, we randomly crop a volume from the whole CT scan with a random size from \{64$\times$64$\times$32, 96$\times$96$\times$64, 96$\times$96$\times$96, 112$\times$112$\times$64\}. Each cropped volume is then resized to 64$\times$64$\times$32. Each voxel's Hounsfield Unit (HU) in the crop is truncated to [-1000,1000]. If a voxel's HU is lower than -150, we regard it as a background voxel. In practice, if over 85\% voxels within a crop belong to the background, we would not use this crop in pre-training.\\

\noindent\textbf{Dataset pre-processing for fine-tuning.} For NIH ChestX-ray and CheXpert, we follow the same pre-processing procedures as in the pre-training stage. On LUNA, we randomly crop a volume for each training iteration, and the size of each crop is 48$\times$48$\times$48. On LiTS, we first localize the liver and expand the target volume by 30 slices on each axis. After random crop, the size of each crop is 256$\times$256$\times$64. Unlike LUNA, we truncate the HU of each voxel to [-200, 200]. For BraTS, the size of each random crop is 112$\times$112$\times$112$\times$4.\\

\noindent\textbf{Data augmentation and multi-crop/sub-crop.} As shown in Fig.~\ref{overview}, there are two types of augmentations, i.e., global and local augmentations. Specifically, for 2D tasks, the global augmentation includes random crop, random horizontal flip, and random rotation. The local augmentation involves random grayscale, gaussian blur, and cutout. In comparison, for 3D tasks, the global augmentation consists of random flip and random affine. Local augmentation strategies are applied, including Gaussian blur, random noise, random gamma, and random swap. Note that all 3D augmentations are implemented following~\cite{perez2021torchio}. As for multi-crop in 2D tasks, we resort to the scale factor of random crop\footnote{\url{https://pytorch.org/vision/main/generated/torchvision.transforms.RandomResizedCrop.html}} to generate global and local views. Specifically, we set the range of scale to [0.3, 1] to generate two global views. For six local views, the scale range is set to [0.05, 0.3]. Both global and local views are resized to 224$\times$224. As for sub-crop in 3D tasks, we randomly sample two global views with a random size from \{64$\times$64$\times$32, 96$\times$96$\times$64, 96$\times$96$\times$96, 112$\times$112$\times$64\}. The IoU constraint (i.e., threshold) between two global views is 0.3. Then, we find the minimum bounding box of global views, from which six local views are randomly cropped, each with a random size from \{8$\times$8$\times$8, 16$\times$16$\times$16, 32$\times$32$\times$16, 32$\times$32$\times$32\}. After random crop, all 3D global views are resized to 64$\times$64$\times$32, while all local views are resized to 16$\times$16$\times$16.\\

\begin{figure}[t]
    \centering
    \includegraphics[width=0.8\columnwidth]{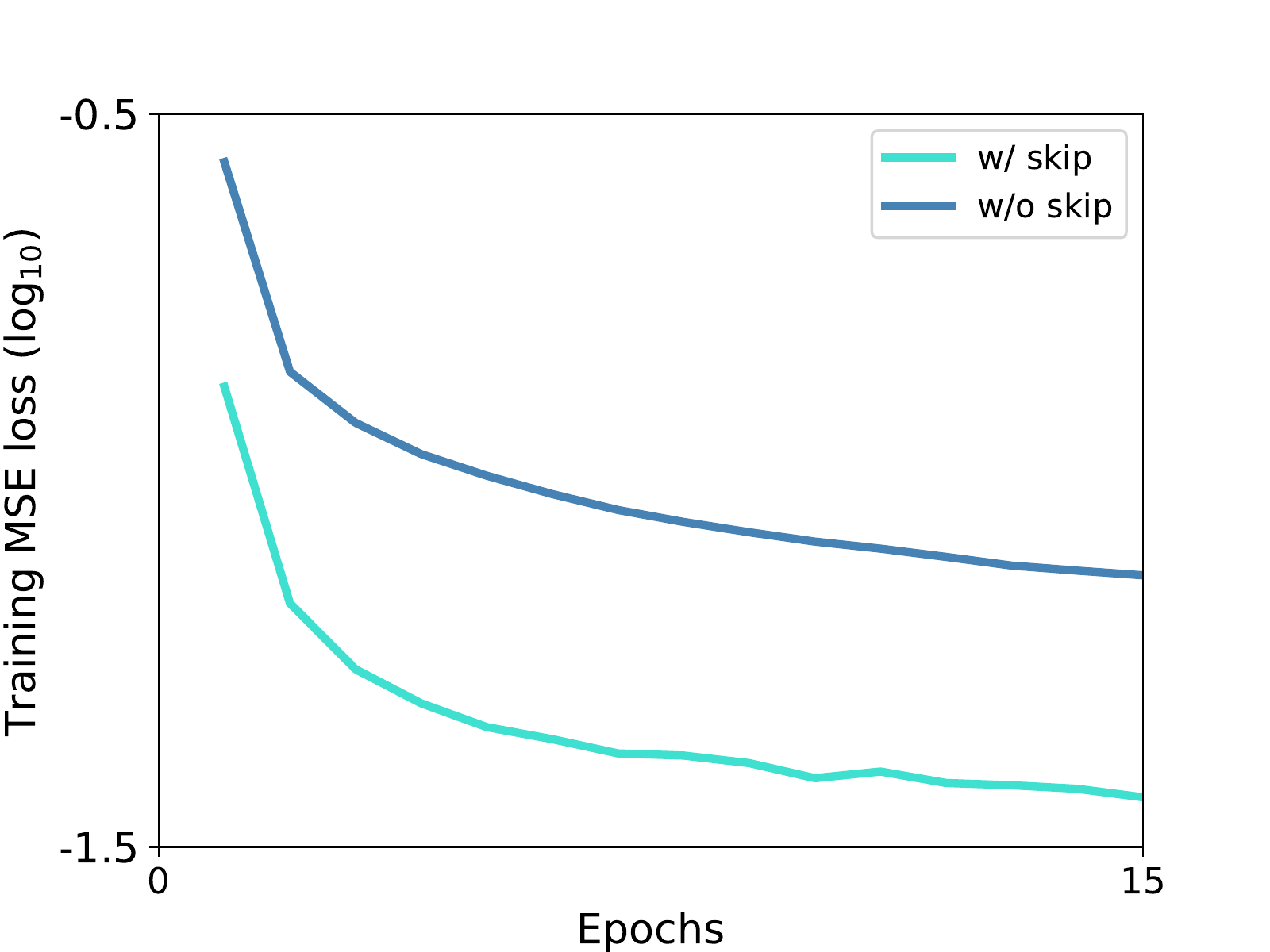}
    \caption{Influence of skip connections in pixel restoration. We display the loss curve of mean square error (MSE) in the first 15 epoches.}
    \label{mse_curve}
\end{figure}
\begin{table}[t]
    \centering
    \caption{Impact of skip connections on chest pathology identification (NIH ChestX-ray), brain tumor segmentation (BraTS), and abdominal organ segmentation (LiTS). On NIH, We use 95\% unlabeled training data for pre-training, while the rest 5\% data with labels are used for fine-tuning. On BraTS and LiTS, we use 10\% labeled data for fine-tuning.}
    \begin{tabular}{|c|c|c|c|}
    \hline
         Datasets &  w/o skip & w/ skip & Gain \\
         \hline
         \hline
         NIH & 76.6 & 75.4 & 1.2 \\
         BraTS & 73.0 & 71.5 & 1.5 \\
         LiTS & 79.0 & 77.6 & 1.4 \\
    \hline
    \end{tabular}
    \label{tab:skip}
\end{table}
\noindent\textbf{Training and evaluation details.} We use stochastic gradient descent (SGD) with momentum as the default optimizer, where the momentum is set to 0.9. The initial learning rate is 1e-2, and we employ the cosine annealing strategy for learning rate decay. We set the weight decay to 1e-5. The number of training epochs is 240. The batch sizes of 2D pre-training and fine-tuning (on NIH ChestX-ray or CheXpert) are 256 and 512, respectively. As for 3D pre-training, the batch size (on LUNA) is 32. For 3D fine-tuning tasks, the batch sizes on LUNA, LiTS, and BraTS are 32, 4, and 4, respectively. The evaluation metric on NIH ChestX-ray, CheXpert, and LUNA is AUROC (Area Under the Receiver Operating Characteristics). For segmentation tasks on LiTS and BraTS, we use Dice similarity as the evaluation metric. We use 70\%, 10\%, and 20\% of the whole dataset to build the training, validation, and test sets. In particular, for semi-supervised learning, we construct the pre-training set by removing a specific amount of data from the entire training set. At the same time, the remainder is used as the training set for fine-tuning. Binary cross-entropy loss is used for the fine-tuning of NIH ChestX-ray, CheXpert, and LUNA, while Dice loss is used for the fine-tuning of LiTS and BraTS.

\subsection{Ablation studies}
\noindent\textbf{Impact of skip connections on pixel restoration.} In Fig.~\ref{mse_curve}, we present the mean square error (MSE) loss (cf. Eq.~\ref{mse_eq}) curves during the training stage. We see that the MSE loss, with skip connections, decreases rapidly in the first 15 training epochs. In comparison, the proposed nsUNet (w/o skip) slows down the decreasing rate of MSE loss. These phenomena are consistent with the role of skip connections, which bridges the gap between low-level pixel details and high-level latent semantics. The existence of skip connections makes it easier to restore pixels by incorporating pixel-level details from low-level but high-resolution feature maps. However, nsUNet removes skip connections, avoiding shortcut solutions to context restoration. Although this design makes it harder to restore pixels (higher loss values in Fig.~\ref{mse_curve}), it helps encode pixel-level information into high-level semantic representations. Such advantage can be verified by the performance gains in Table~\ref{tab:skip}, where removing skip connections brings over 1\% improvement to chest pathology identification, brain tumor segmentation, and abdominal organ segmentation.\\

\begin{figure}[t]
    \centering
    \subfloat[][Pairwise]{\includegraphics[width=0.45\columnwidth]{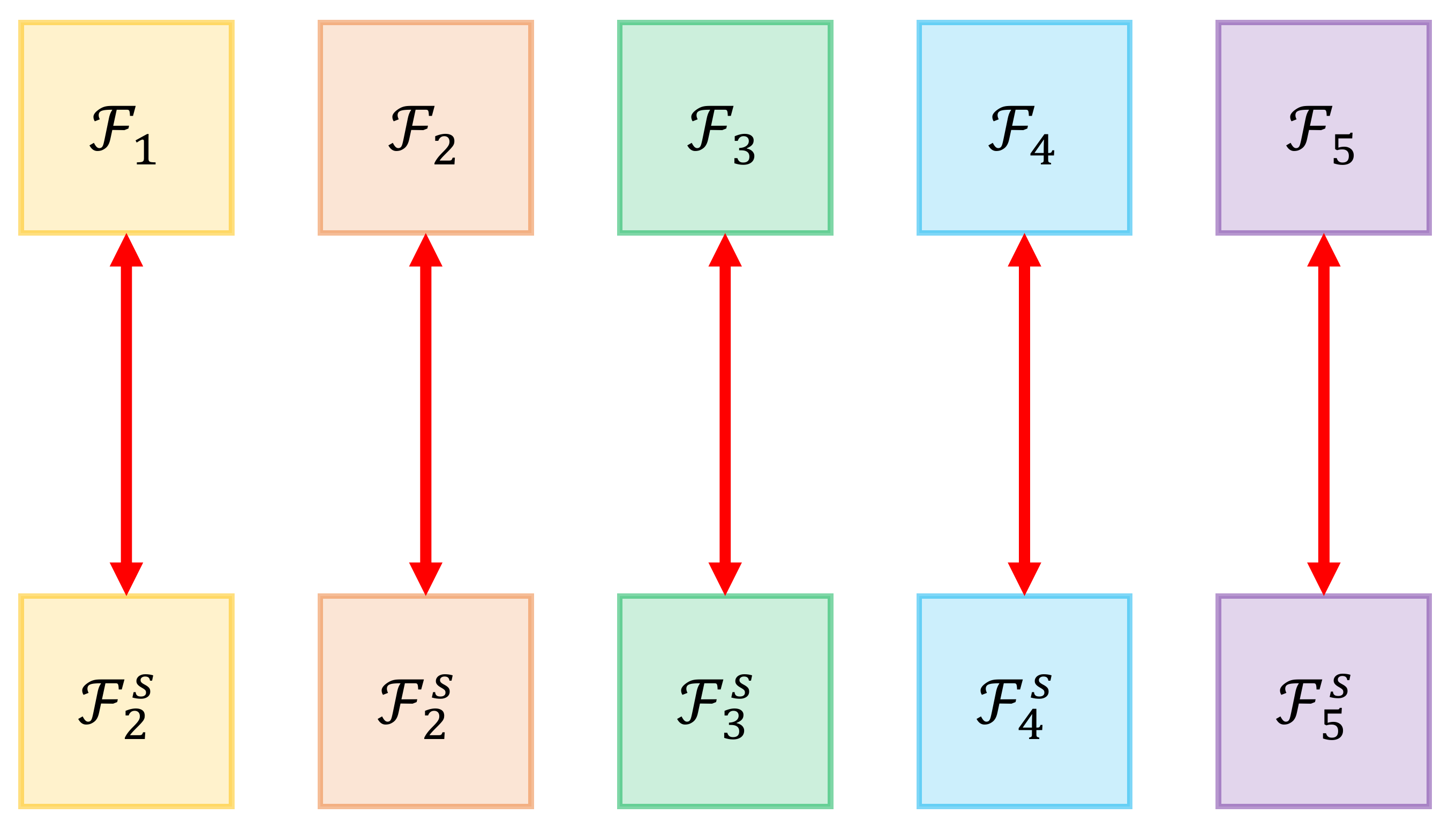}\label{pairwise}}\quad\quad
    \subfloat[][Cross-scale]{\includegraphics[width=0.45\columnwidth]{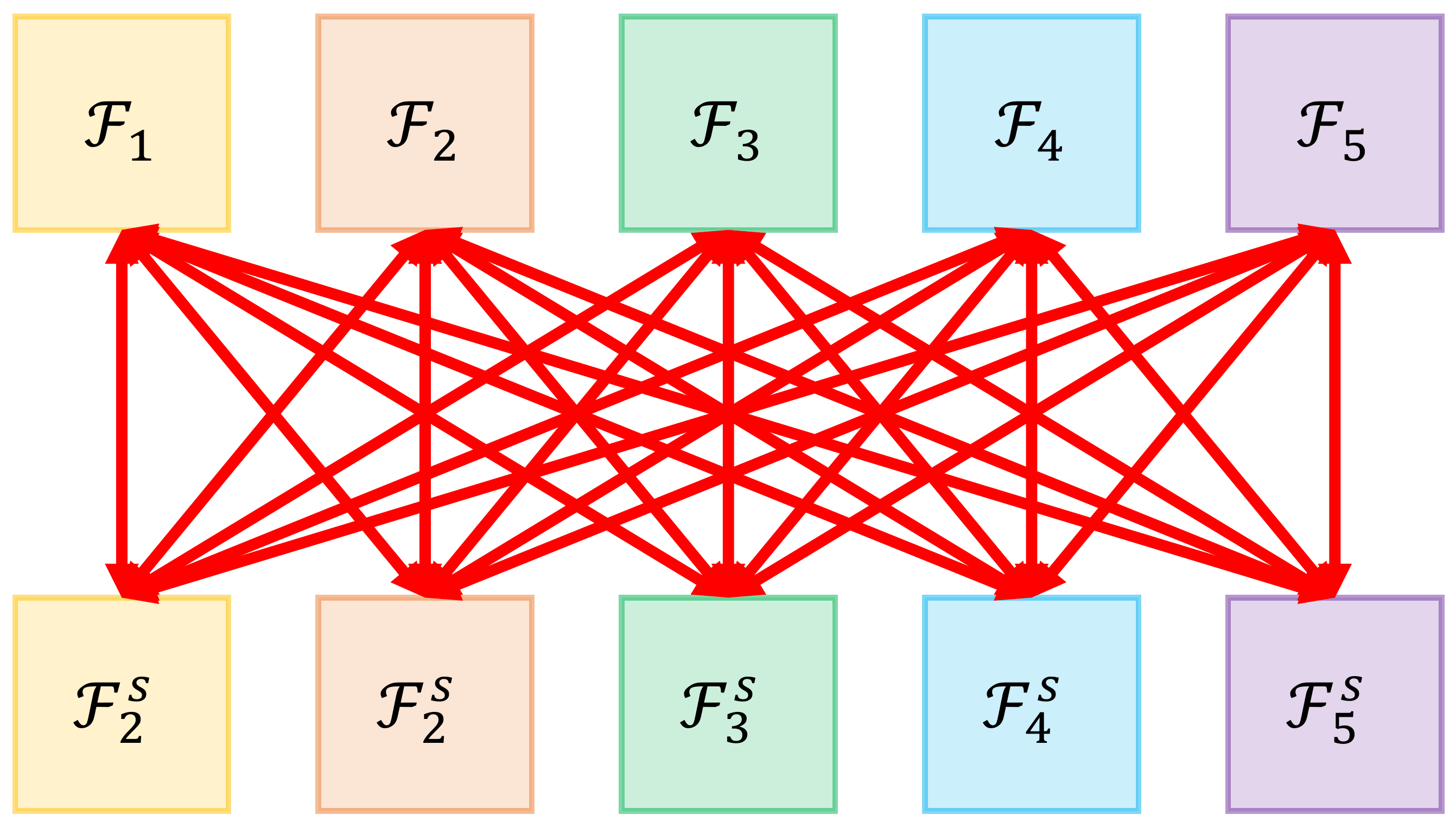}\label{crossed}}    
    \caption{Two choices of how to conduct siamese feature comparison for multiple feature scales. Here, we primarily consider pairwise feature comparison and cross-scale feature comparison.}
    \label{scale_choice}
\end{figure}

\noindent\textbf{How to conduct siamese feature comparison for multiple feature scales?} We illustrate two intuitive choices in Fig.~\ref{scale_choice}. Besides the adopted pairwise comparison manner (Fig.~\ref{pairwise}), another obvious choice is to compare siamese features following a crossed way (a similar strategy was used in~\cite{bachman2019learning}). As shown in Fig.~\ref{crossed}, the cross-scale comparison aggressively compares siamese features across all feature scales. The motivation behind is to introduce multi-scale latent representations by coupling features across different scales. Table~\ref{tab:scales} reports the experimental results of pairwise and cross-scale siamese feature comparison. We find that cross-scale feature comparison slightly deteriorates the performance of semi-supervised pathology identification by 0.6 percents. The underlying reason might be that the features in each scale maintains distinct characteristics, and neglecting these discrepancies can lead to degenerate feature representations.\\

\noindent\textbf{Investigation of different modules in PCRLv2.} In Table~\ref{tab:ab}, we study and report the impact of different modules on the whole tumor (WT) and enhancing tumor (ET) classes of BraTS. Note that in practice, most instances of WT are much larger than instances from ET, making ET instances harder to segment. Besides, we also present the transfer learning results on NIH ChestX-ray.

\begin{table}[!htp]
    \centering
    \caption{Results of pairwise and crossed siamese feature comparison (semi-supervised learning on NIH ChestX-ray). The ratio of unlabeled to labeled data is 9.5:0.5.}
    \begin{tabular}{|c|c|c|c|}
    \hline
         & Pairwise & Crossed~\cite{bachman2019learning} & Gain \\
         \hline
         \hline
         Mean AUROC& 76.6 & 76.0 & 0.6\\
    \hline
    \end{tabular}
    \label{tab:scales}
\end{table}
\begin{table*}[!htp]
    \centering
    \caption{Impact of different modules in PCRLv2. \textbf{Res.} and \textbf{Comp.} denote the tasks of pixel restoration and feature comparison, respectively. \textbf{S (N)} means there are N scales included. \textbf{MC} and \textbf{SC} stand for the multi-crop and proposed sub-crop strategies, respectively. \textbf{WT} and \textbf{ET} denote classes of the whole tumor and enhancing tumor in BraTS, respectively. In most cases, instances from WT are much larger (in size) than those of ET. We performed these experiments by first using LUNA for self-supervised pre-training, and then we fine-tune the pre-trained model on BraTS using 10\% labeled data. \textbf{NIH} denotes the transfer learning on chest pathology identification, where we use CheXpert for pre-training and fine-tune the pre-trained model with 50\% labeled data from NIH ChestX-ray.}
    \resizebox{1.5\columnwidth}{!}{
    \begin{tabular}{|c|c|c|c|c|c|c|c|c|c|}
    \hline
         {\textcolor{gray}{\#}} & Res. & Comp. & S (3) & S (5) & MC & SC & WT (BraTS) & ET (BraTS) & NIH \\
         \hline
         \hline
         {\textcolor{gray}0} & \checkmark & & & & & &74.2 &64.9 & 78.2 \\
         {\textcolor{gray}1} & & \checkmark & & & & & 76.4 & 63.8 & 78.5 \\
         {\textcolor{gray}2} & \checkmark & \checkmark & & & & &76.2 &64.6 & 80.9\\
         {\textcolor{gray}3} & \checkmark & \checkmark &\checkmark & & & & 76.9 & 66.1 & 81.5\\
         {\textcolor{gray}4} & \checkmark & \checkmark & &\checkmark & & & 77.2 & 66.8 & 82.0\\
         {\textcolor{gray}5} & \checkmark & \checkmark & &\checkmark &\checkmark & & \emph{fail} & \emph{fail} & 82.5\\
         {\textcolor{gray}6}& \checkmark & \checkmark & &\checkmark & & \checkmark & 77.7 & 67.2 & 82.7\\
    \hline
    \end{tabular}}
    \label{tab:ab}
\end{table*}

First of all, we investigate the influence of pixel restoration (row 0) and feature comparison (row 1), respectively. We directly reconstruct the full resolution uncorrupted images for the pixel restoration task while siamese feature comparison is conducted on the last-layer output of the encoder. Comparing row 0 with row 1, we see that the context restoration task is more advantageous in segmentation of small tumor regions (i.e., ET)  while the comparative SSL is more capable of dealing with large tumor regions (i.e., WT) and chest pathologies. Such comparison shows that semantic information preservation may be more helpful to the detection of large objects, while segmenting small objects requires the incorporation of pixel-level information. In row 2, we can already acquire noticeable performance gains by directly combining pixel restoration and feature comparison.

\begin{table*}[!htp]
    \centering
    \caption{Semi-supervised chest pathology identification (on NIH ChestX-ray). The labeling ratio denotes the amount of data with labels in the training set that is used for fine-tuning while the remaining data in the training set is used for self-supervised pre-training. The best results are bolded.}
    \resizebox{0.9\textwidth}{!}{
    \begin{tabular}{|c|c|c|cccccccccccccc|}
    \hline
         \rotatebox{90}{Labeling ratio} & \rotatebox{90}{Methodology} & \rotatebox{90}{Mean} & \rotatebox{90}{Atelectasis} & \rotatebox{90}{Cardiomegaly} & \rotatebox{90}{Effusion} & \rotatebox{90}{Infiltration} & \rotatebox{90}{Mass} & \rotatebox{90}{Nodule} & \rotatebox{90}{Pneumonia} & \rotatebox{90}{Pneumothorax} & \rotatebox{90}{Consolidation} & \rotatebox{90}{Edema} & \rotatebox{90}{Emphysema} & \rotatebox{90}{Fibrosis} & \rotatebox{90}{Pleural Thick.} & \rotatebox{90}{Hernia}  \\
    \hline
    \hline
    \multirow{7}{*}{5\%}& TS &61.8& 58.8& 72.0&  68.8& 51.5& 63.8& 49.2& 57.4& 67.4& 61.5& 71.0&  62.7& 58.1& 60.0& 63.1\\ 
    & MG~\cite{zhou2021models} & 66.4& 63.4& 74.1& 72.9& 53.5& 67.2& 54.3& 59.9& 71.3& 66.5& 77.0&  65.8& 64.5& 62.8& 76.2\\ 
    & TransVW~\cite{haghighi2021transferable} &66.5& 64.2& 72.9& 72.2& 54.8& 69.4& 55.7& 59.6& 71.0&  64.8& 77.4& 66.6& 63.6& 62.8& 75.6\\ 
    & C2L~\cite{zhou2020comparing}&	71.7& 69.9& 77.9& 76.2& 59.1& 73.4& 60.0&  64.5& 76.2& 71.4& 80.3& 76.1& 69.9& 68.4& 80.4\\
    & SimSiam~\cite{chen2021exploring}&	71.7& 68.9& 79.3& 77.8& 58.7& 73.0&  61.0&  65.4& 76.2& 72.1& 81.7& 75.1& 69.6& 68.1& 76.8\\ \cline{2-17}
    & PCRLv1~\cite{zhou2021preservational} & 74.1 & 70.1& 80.3& 79.3& 61.8& \textbf{76.8}& 64.6& 68.6& 77.2& 72.8& 83.7& 77.4& 71.3& 72.7& 80.8\\
    & PCRLv2 & \textbf{76.6}& \textbf{75.7}& \textbf{81.0}& \textbf{80.3}& \textbf{64.0}& \textbf{76.8}& \textbf{68.7}& \textbf{70.7}& \textbf{83.2}& \textbf{77.5}& \textbf{87.8}& \textbf{79.2}& \textbf{72.5}& \textbf{73.2}& \textbf{81.8}\\
    \hline
    \multirow{7}{*}{10\%}& TS &68.1& 65.8& 77.6& 74.4& 57.1& 69.4& 54.8& 63.0&  72.9& 68.3& 78.8& 68.2& 64.3& 66.4& 72.5\\ 
    & MG~\cite{zhou2021models} & 70.0 & 67.1 & 78.1 & 76.1 & 57.2 & 72.8 & 57.5& 63.3& 75.5& 70.9& 79.5& 68.8& 67.4& 68.0& 77.6\\ 
    & TransVW~\cite{haghighi2021transferable} &70.2& 66.6& 78.9& 74.9& 58.4& 71.2& 59.5& 64.8& 72.6& 70.4& 79.4& 70.7& 67.2& 68.3& 79.5\\ 
    & C2L~\cite{zhou2020comparing}&	74.1& 72.3& 81.7& 79.9& 60.2& 74.6& 62.7& 67.6& 78.7& 73.9& 83.5& 78.2& 72.8& 69.8& 81.4\\
    & SimSiam~\cite{chen2021exploring}&	74.0& 71.2& 81.4& 78.9& 60.2& 75.5& 63.2& 67.3& 78.7& 73.2& 83.5& 77.7& 72.5& 71.8& 80.8\\ \cline{2-17}
    & PCRLv1~\cite{zhou2021preservational} & 76.2& 73.6& 82.9& 81.2& 64.7 &77.1& 66.7& 69.7& 79.8& 74.5& 86.9& 78.8& 75.6& \textbf{74.2}& 81.1\\
    & PCRLv2 & \textbf{78.2}& \textbf{77.2}& \textbf{84.3}& \textbf{84.4}& \textbf{67.4}& \textbf{77.5}& \textbf{68.9}& \textbf{71.6}& \textbf{84.4}& \textbf{77.8}& \textbf{89.0}& \textbf{79.3}& \textbf{76.1}& 74.0& \textbf{82.4}\\
    \hline
    \multirow{7}{*}{20\%}& TS &71.5& 68.9& 80.7& 77.5& 60.2& 73.6& 58.7& 66.2& 76.1& 71.7& 82.9& 72.2& 69.0& 68.7& 74.7\\ 
    & MG~\cite{zhou2021models} & 73.9& 71.9& 83.0&  80.0&  62.3& 75.2& 62.2& 67.5& 79.0&  73.3& 83.6& 73.4& 71.0&  70.6& 81.4\\ 
    & TransVW~\cite{haghighi2021transferable} &74.3& 71.6& 82.5& 80.1& 62.3& 76.7& 62.8& 69.2& 78.2& 73.5& 83.8& 75.4& 72.2& 71.2& 80.3\\ 
    & C2L~\cite{zhou2020comparing}&	76.4& 74.2& 83.9& 81.7& 63.8& 77.3& 64.7& 70.3& 81.5& 75.5& 86.0&  80.2& 75.2& 73.4& 81.8\\
    & SimSiam~\cite{chen2021exploring}&	76.5& 73.8& 84.0&  81.4& 63.2& 78.2& 64.7& 69.6& 82.1& 76.2& 86.4& 80.7& 75.0&  73.9& 81.7 \\ \cline{2-17}
    & PCRLv1~\cite{zhou2021preservational} & 78.8& 75.4& 86.2& 83.6& 65.1& 79.9& 69.6& 72.0&  82.3& 79.9& 88.3& 82.6& 76.5& 75.9& 81.9\\
    & PCRLv2 & \textbf{79.9}& \textbf{78.1}& \textbf{87.2}& \textbf{85.9}& \textbf{68.2}& \textbf{80.5}& \textbf{69.9}& \textbf{72.5}& \textbf{85.3}& \textbf{80.4}& \textbf{89.2}& \textbf{83.1}& \textbf{77.5}& \textbf{77.0}& \textbf{83.5}\\
    \hline
    \multirow{7}{*}{30\%}& TS &73.4& 70.6& 81.9& 79.1& 61.6& 75.5& 60.7& 68.8& 78.3& 72.7& 84.3& 74.1& 70.3& 70.9& 78.9\\ 
    & MG~\cite{zhou2021models} & 76.1& 74.3& 84.4& 82.1& 63.6& 78.3& 64.4& 69.6& 81.2& 75.8& 85.6& 75.9& 73.6& 73.6& 82.8\\ 
    & TransVW~\cite{haghighi2021transferable} &76.7& 74.9& 84.1& 81.9& 64.9& 79.0&  65.3& 70.9& 80.3& 76.2& 86.5& 78.6& 74.5& 74.2& 82.1\\ 
    & C2L~\cite{zhou2020comparing}&	77.5& 74.3& 84.8& 82.6& 64.6& 78.3& 66.3& 71.5& 83.0&  76.8& 87.6& 81.3& 76.5& 74.4& 82.9\\
    & SimSiam~\cite{chen2021exploring}&	78.0&  75.4& 85.1& 82.9& 65.0&  79.4& 67.0&  71.4& 83.4& 77.4& 87.8& 82.8& 76.1& 75.5& 82.7 \\ \cline{2-17}
    & PCRLv1~\cite{zhou2021preservational} & 79.0&  75.5& 86.6& 83.8& 65.9& 80.7& 70.2& 72.8& 82.9& 80.4& 88.9& 83.3& 76.6& 76.5& 81.9\\
    & PCRLv2 & \textbf{81.1}& \textbf{78.4}& \textbf{87.6}& \textbf{86.6}& \textbf{69.6}& \textbf{82.8}& \textbf{72.0}& \textbf{74.0}& \textbf{86.2}& \textbf{81.0}& \textbf{89.9}& \textbf{84.4}& \textbf{79.5}& \textbf{79.0}& \textbf{84.6}\\   
    \hline
    \multirow{7}{*}{40\%}& TS &75.4& 72.6& 83.6& 81.5& 62.9& 77.3& 63.3& 70.1& 80.3& 74.9& 85.5& 76.4& 72.5& 73.0& 81.8\\ 
    & MG~\cite{zhou2021models} & 77.3& 75.4& 86.0&  83.3& 65.1& 79.0& 65.1& 70.8& 82.1& 77.0&  87.3& 76.7& 74.8& 74.9& 83.5\\ 
    & TransVW~\cite{haghighi2021transferable} &77.6& 75.0&  85.1& 82.7& 65.2& 79.7& 66.5& 72.0&  81.0&  76.7& 87.2& 79.2& 75.5& 76.5& 83.7\\ 
    & C2L~\cite{zhou2020comparing}&	79.0&  76.0&  86.1& 84.3& 66.0&  80.0&  67.9& 72.5& 84.1& 78.5& 88.5& 83.7& 77.9& 76.6& 83.8\\
    & SimSiam~\cite{chen2021exploring}&	79.4& 76.7& 86.7& 84.7& 67.0&  80.9& 69.0&  73.1& 84.4& 78.9& 88.9& 83.5& 77.7& 76.6& 83.4 \\ \cline{2-17}
    & PCRLv1~\cite{zhou2021preservational} & 79.9& 76.7& 87.1& 84.9& 67.1& 82.7& 72.2& 73.3& 83.6& 80.6& 89.2& 83.8& 77.3& 76.9& 83.2\\
    & PCRLv2 & \textbf{81.5}& \textbf{78.7}& \textbf{87.8}& \textbf{87.0}& \textbf{69.8}& \textbf{83.2}& \textbf{72.5}& \textbf{74.7}& \textbf{86.3}& \textbf{81.2}& \textbf{90.2}& \textbf{84.9}& \textbf{80.0}& \textbf{79.4}& \textbf{85.0}\\ 
    \hline
    \end{tabular}}
    \label{ssl:chest14}
\end{table*}
\begin{table}[!htp]
    \centering
    \caption{Semi-supervised pulmonary nodule detection (on LUNA). The labeling ratio indicates how much data from the training set with labels is utilized for fine-tuning while the rest of the data is used for pre-training. Best results are bolded.}
    \resizebox{0.8\columnwidth}{!}{
    \begin{tabular}{|c|c|c|c|c|}
    \hline
         \multirow{2}{*}{Methodology} & \multicolumn{4}{c|}{Labeling ratio}\\ \cline{2-5}
         & 10\% & 20\% & 30\% & 40\%  \\
         \hline
         \hline
         TS & 78.4 & 83.0 & 85.7 & 87.5\\
         MG~\cite{zhou2021models} & 80.2& 85.0& 87.5& 90.3\\
         TransVW~\cite{haghighi2021transferable} &79.3& 84.5& 87.9& 90.5\\
         Cube++~\cite{tao2020revisiting} &81.4& 85.2& 87.9& 90.0\\
         3D-CPC~\cite{taleb20203d}&80.2& 85.2& 88.3& 90.6\\
         \hline
         PCRLv1~\cite{zhou2021preservational} & 84.4 & 87.5& 89.8& 92.2\\
         PCRLv2 & \textbf{85.5} & \textbf{88.3}& \textbf{90.3}& \textbf{93.1}\\
    \hline
    \end{tabular}}
    \label{ssl:luna}
\end{table}

Next, we show that multi-scale representations benefit both pixel restoration and feature comparison tasks. By conducting both tasks on 3 scales, we observe a 0.7-percent improvement on WT, a 1.5-percent gain on ET, and a 0.6-percent improvement on chest pathology classification. These results show that introducing multiple scales is more helpful to the segmentation of small regions. Moreover, by increasing the number of scales from 3 to 5, we can improve the accuracy of all three tasks consistently. Not surprisingly, ET benefits the most from the introduction of multiple scales, indicating the necessity of utilizing multi-scale representations in medical image segmentation.

\begin{table*}[!htp]
    \centering
    \caption{Transfer learning on chest pathology identification. We pre-train the model using data from CheXpert (without labels). Then, we fine-tune the pre-trained model on NIH ChestX-ray with different amounts of labeled data (denotes as different labeling ratios). The best results are bolded.}
    \resizebox{0.9\textwidth}{!}{
    \begin{tabular}{|c|c|c|cccccccccccccc|}
    \hline
         \rotatebox{90}{Labeling ratio} & \rotatebox{90}{Methodology} & \rotatebox{90}{Mean} & \rotatebox{90}{Atelectasis} & \rotatebox{90}{Cardiomegaly} & \rotatebox{90}{Effusion} & \rotatebox{90}{Infiltration} & \rotatebox{90}{Mass} & \rotatebox{90}{Nodule} & \rotatebox{90}{Pneumonia} & \rotatebox{90}{Pneumothorax} & \rotatebox{90}{Consolidation} & \rotatebox{90}{Edema} & \rotatebox{90}{Emphysema} & \rotatebox{90}{Fibrosis} & \rotatebox{90}{Pleural Thick.} & \rotatebox{90}{Hernia}  \\
    \hline
    \hline
    \multirow{8}{*}{10\%}& TS &68.1& 67.6& 63.3& 76.8& 57.5& 71.5& 61.8& 64.2& 76.2& 69.8& 80.2& 72.4& 62.8& 68.0& 61.1\\ 
    & IN~\cite{oord2018representation} & 73.5& 73.3& 68.7& 81.6& 63.0&  76.6& 67.3& 70.0&  81.3& 75.6& 85.9& 78.5& 68.6& 72.5& 65.9 \\
    & MG~\cite{zhou2021models} & 70.1& 69.9& 65.6& 79.2& 59.4& 72.9& 64.3& 67.0&  77.9& 72.0&  82.3& 75.8& 65.9& 69.6& 59.4\\ 
    & TransVW~\cite{haghighi2021transferable} &69.7& 69.4& 64.3& 78.2& 59.5& 72.6& 63.1& 67.2& 77.2& 70.9& 83.0&  75.3& 65.8& 68.9& 60.2\\ 
    & C2L~\cite{zhou2020comparing}&	73.1& 72.5& 68.0&  81.3& 62.4& 75.8& 67.2& 70.2& 80.6& 74.8& 85.4& 78.4& 68.3& 72.2& 66.1\\
    & SimSiam~\cite{chen2021exploring}&	72.5& 71.9& 67.5& 81.2& 61.7& 75.9& 66.6& 69.6& 79.8& 74.2& 84.8& 77.6& 67.7& 71.8& 64.5\\ \cline{2-17}
    & PCRLv1~\cite{zhou2021preservational} & 75.8& 75.4& 70.6& 84.2& 65.5& 78.9& 69.6& 72.7& 83.5& 77.6& 88.5& 80.8& 71.3& 74.8& 67.6\\
    & PCRLv2 & \textbf{77.2} & \textbf{76.8}& \textbf{72.0}& \textbf{85.6}& \textbf{66.8}& \textbf{80.2}& \textbf{71.0}& \textbf{74.0}& \textbf{84.8}& \textbf{78.9}& \textbf{89.8}& \textbf{82.2}& \textbf{72.6}& \textbf{76.2}& \textbf{69.7}\\
    \hline
    \multirow{8}{*}{20\%}& TS &71.4& 71.8& 73.1& 78.4& 59.6& 72.5& 64.5& 66.6& 77.7& 71.7& 82.0&  75.5& 69.8& 68.9& 68.2\\
    & IN~\cite{deng2009imagenet} & 76.2& 75.9& 78.3& 82.9& 64.2& 77.8& 68.8& 70.7& 83.0&  76.4& 87.2& 80.0&  75.3& 73.9& 73.1\\
    & MG~\cite{zhou2021models} & 73.8& 73.9& 75.4& 80.2& 61.9& 74.9& 66.5& 68.3& 80.0&  74.0&  85.1& 78.1& 72.8& 71.5& 71.3\\ 
    & TransVW~\cite{haghighi2021transferable} &73.8& 73.0&  75.5& 80.1& 62.3& 75.6& 66.7& 68.6& 80.2& 74.0&  85.2& 77.5& 72.9& 71.5& 69.4\\ 
    & C2L~\cite{zhou2020comparing}&	77.0&  76.5& 78.9& 83.4& 65.0&  78.6& 69.8& 71.8& 83.5& 77.2& 88.1& 80.8& 76.0&  74.2& 73.5\\
    & SimSiam~\cite{chen2021exploring}&	76.6& 76.6& 78.7& 83.3& 64.6& 77.9& 69.2& 71.6& 83.1& 76.9& 87.8& 80.5& 75.5& 73.8& 73.6\\ \cline{2-17}
    & PCRLv1~\cite{zhou2021preservational} & 77.5& 77.3& 79.7& 84.3& 65.7& 78.9& 70.3& 72.8& 83.8& 77.6& 88.6& 81.1& 76.5& 74.8& 74.3\\
    & PCRLv2 & \textbf{79.4}& \textbf{79.0}& \textbf{81.3}& \textbf{85.9}& \textbf{67.3}& \textbf{80.8}& \textbf{72.1}& \textbf{74.0}& \textbf{86.0}& \textbf{79.4}& \textbf{90.3}& \textbf{83.1}& \textbf{78.4}& \textbf{76.7}& \textbf{76.6}\\
    \hline
    \multirow{8}{*}{30\%}& TS &73.5& 71.7& 79.7& 79.9& 60.5& 76.5& 68.4& 66.8& 79.2& 72.8& 83.4& 76.9& 71.4& 70.5& 71.3\\ 
    & IN~\cite{deng2009imagenet} & 78.5& 77.2& 84.6& 84.3& 66.2& 80.8& 73.0&  72.3& 84.0&  78.0&  88.5& 82.0&  76.8& 75.3& 76.0\\
    & MG~\cite{zhou2021models} & 75.6& 74.1& 81.8& 81.0&  63.3& 77.9& 70.1& 69.0&  80.9& 74.8& 85.4& 79.7& 73.6& 72.6& 74.2\\ 
    & TransVW~\cite{haghighi2021transferable} &75.7& 74.8& 81.4& 81.0&  63.6& 77.7& 69.9& 69.8& 80.9& 75.4& 86.0&  79.3& 73.9& 72.3& 73.8\\ 
    & C2L~\cite{zhou2020comparing}&	78.6& 77.1& 84.5& 84.5& 66.1& 81.1& 73.0&  72.5& 84.0&  78.1& 88.3& 82.1& 76.8& 75.5& 76.8\\
    & SimSiam~\cite{chen2021exploring}&	78.3& 77.0&  84.4& 84.1& 65.7& 80.7& 72.7& 72.2& 83.9& 77.9& 88.1& 82.1& 76.6& 75.2& 75.6 \\ \cline{2-17}
    & PCRLv1~\cite{zhou2021preservational} & 79.9& 78.5& 85.8& 85.6& 67.4& 82.3& 74.2& 73.8& 85.5& 79.4& 89.7& 83.5& 78.1& 76.7& 78.1\\
    & PCRLv2 & \textbf{80.5} & \textbf{79.1}& \textbf{86.4}& \textbf{86.2}& \textbf{68.0}& \textbf{82.8}& \textbf{74.8}& \textbf{74.3}& \textbf{86.0}& \textbf{80.0}& \textbf{90.3}& \textbf{84.1}& \textbf{78.6}& \textbf{77.2}& \textbf{79.2}\\
    \hline
    \multirow{8}{*}{40\%}& TS & 75.4& 72.6& 80.0&  81.0&  62.5& 76.9& 69.2& 68.0&  80.7& 74.7& 85.1& 79.5& 74.0&  71.0& 79.8\\ 
    & IN~\cite{deng2009imagenet} & 79.0&  76.7& 84.2& 84.3& 66.3& 80.7& 73.6& 72.3& 84.7& 78.5& 88.6& 83.4& 77.4& 75.0& 79.7\\
    & MG~\cite{zhou2021models} & 76.5& 74.1& 81.3& 81.7& 63.9& 77.9& 71.1& 70.1& 82.5& 76.1& 85.6& 80.6& 74.5& 73.1& 77.9\\ 
    & TransVW~\cite{haghighi2021transferable} &77.3& 75.2& 82.4& 82.4& 64.4& 79.0&  71.4& 70.5& 83.2& 76.7& 86.6& 82.0&  75.8& 73.6& 78.4\\ 
    & C2L~\cite{zhou2020comparing}&	79.1& 76.9& 84.3& 84.5& 66.4& 80.8& 73.4& 72.2& 84.8& 78.3& 88.6& 83.4& 77.2& 75.4& 80.6\\
    & SimSiam~\cite{chen2021exploring}&	78.9& 76.7& 83.9& 84.1& 66.6& 80.4& 73.1& 72.1& 84.7& 78.1& 88.4& 83.4& 77.2& 74.8& 80.5 \\ \cline{2-17}
    & PCRLv1~\cite{zhou2021preservational} & 80.8& 78.5& 86.0&  86.2& 68.2 &82.4& 75.2& 74.0&  86.6& 80.2& 90.2& 85.1& 79.0&  76.9& 82.1\\
    & PCRLv2 & \textbf{81.5} & \textbf{79.2}& \textbf{86.6}& \textbf{86.9}& \textbf{68.9}& \textbf{83.0}& \textbf{75.8}& \textbf{74.6}& \textbf{87.2}& \textbf{80.8}& \textbf{90.9}& \textbf{85.8}& \textbf{79.7}& \textbf{77.6}& \textbf{83.4}\\   
    \hline
    \multirow{8}{*}{50\%}& TS &77.5& 75.2& 82.0&  82.0&  64.5& 79.6& 71.8& 71.3& 82.9& 75.8& 86.6& 80.9& 76.1& 75.5& 80.3\\ 
    & IN & 79.5& 77.2& 84.5& 84.4& 66.6& 81.4& 73.6& 73.0&  84.6& 78.2& 89.1& 82.7& 77.9& 77.3& 82.0\\
    & MG~\cite{zhou2021models} & 77.6& 75.0&  82.8& 82.8& 64.8& 79.5& 71.8& 71.6& 82.3& 75.7& 86.7& 81.5& 76.2& 75.7& 79.5\\ 
    & TransVW~\cite{haghighi2021transferable} &77.3& 74.5& 81.9& 82.4& 64.8& 78.8& 71.5& 71.3& 82.4& 75.7& 86.8& 80.4& 75.7& 74.9& 80.6\\ 
    & C2L~\cite{zhou2020comparing}&	79.8& 77.6& 84.7& 84.5& 67.0&  81.6& 73.6& 73.4& 84.7& 78.5& 89.0&  83.1& 78.4& 78.0&82.6\\
    & SimSiam~\cite{chen2021exploring}&	80.0 & 77.7 & 84.9& 84.8& 67.1& 81.7& 74.0&  73.5& 84.7& 78.3& 89.5& 83.6& 78.8& 77.7& 83.2 \\ \cline{2-17}
    & PCRLv1~\cite{zhou2021preservational} & 81.2& 78.7& 86.1& 86.3& 68.3& 82.8& 75.4& 74.5& 86.8& 80.4& 90.5& 85.3& 79.5& 78.2& 83.5\\
    & PCRLv2 & \textbf{82.5}& \textbf{80.0}& \textbf{87.4}& \textbf{87.3}& \textbf{69.6}& \textbf{84.1}& \textbf{76.4}& \textbf{76.1}& \textbf{87.4}& \textbf{81.0}& \textbf{91.8}& \textbf{85.9}& \textbf{81.0}& \textbf{80.4}& \textbf{86.1}\\ 
    \hline
    \multirow{8}{*}{100\%}& TS &80.9& 77.7& 86.1& 85.1& 67.7& 84.2& 73.3& 73.9& 84.9& 78.7& 89.4& 85.4& 79.4& 78.5& 87.6\\ 
    & IN & 80.8& 77.8& 86.3& 84.7& 67.3& 83.6& 73.0&  74.1& 84.9& 78.8& 89.5& 85.7& 79.6& 78.2& 87.0\\
    & MG~\cite{zhou2021models} & 80.8& 77.8& 86.3& 84.7& 67.3& 83.6& 73.0&  74.1& 84.9& 78.8& 89.5& 85.7& 79.6& 78.2& 87.0\\ 
    & TransVW~\cite{haghighi2021transferable} &81.2& 77.9& 86.4& 85.3& 67.6& 84.3& 73.8& 74.4& 85.1& 79.3& 89.8& 86.2& 80.0&  78.6& 88.8\\ 
    & C2L~\cite{zhou2020comparing}&	81.4& 78.2& 87.0&  85.3& 68.3& 84.8& 73.7& 74.8& 85.5& 79.6& 90.1& 86.3& 80.0&  78.6& 88.1\\
    & SimSiam~\cite{chen2021exploring}&	81.6& 78.3& 87.2& 85.5& 68.3& 84.9& 74.2& 74.7& 85.7& 79.6& 90.1& 86.2& 80.2& 79.1& 89.1 \\ \cline{2-17}
    & PCRLv1~\cite{zhou2021preservational} & 83.0&  79.8& 88.5& 87.1& 69.7& 86.1& 75.6& 76.1& 87.0&  81.2& 91.6& 87.7& 81.7& 80.4& 90.2\\
    & PCRLv2 & \textbf{84.0}& \textbf{80.7}& \textbf{89.3}& \textbf{87.9}& \textbf{70.5}& \textbf{87.0}& \textbf{76.4}& \textbf{77.0}& \textbf{87.9}& \textbf{82.0}& \textbf{92.5}& \textbf{88.6}& \textbf{82.6}& \textbf{81.3}& \textbf{91.6}\\     
    \hline
    \end{tabular}}
    \label{tl:chest14}
\end{table*}

Last but not the least, we investigate the significance of multi-crop (row 4) and sub-crop (row 5). We empirically found that directly applying multi-crop to 3D medical volumes leads to the failure of model training. The underlying reason might be that it is difficult for cropped global and local views to maintain clear spatial relations in the 3D space as in the 2D space. In contrast, sub-crop can provide consistent performance gains on both types of tumor regions by successfully preserving the spatial relations in latent representations. When applying sub-crop to 2D X-rays, we observe a marginal improvement over multi-crop. The underlying reason is that sub-crop is proposed to handle dispersed sampled views in a 3D space to guarantee the model captures local-global relations. However, in a 2D space, the sampled views usually (partly) overlap. 

\subsection{Semi-supervised chest pathology identification}
Table~\ref{ssl:chest14} presents the experimental results of applying semi-supervised learning on NIH ChestX-ray. Specifically, we use a specific amount of the training set (denoted as the labeling ratio in Table~\ref{ssl:chest14}) as labeled data while the remaining training data is used for self-supervised pre-training.

From Table~\ref{ssl:chest14}, we see that self-supervised pre-training can dramatically boost the performance compared to train from scratch (TS), which verify the necessity of conducting pre-training in medical imaging. Comparing MG with TransVW, they show similar performance in different labeling ratios. Such comparison is easy to explain as TransVW is built upon MG, and both are based on context restoration. TransVW performs slightly better than MG, as it incorporates an additional classification head to encode more semantics. Compared to context restoration based methods, comparative methodologies (C2L and SimSiam) display better overall and class-specific results, especially in small labeling ratios. The underlying reason might be that semantic information is more critical than pixel-level information in chest pathology detection. As for C2L and SimSiam, C2L performs better when the amount of labeled data is quite limited. However, SimSiam gradually produces better diagnosis results as the labeling ratio increases.

After incorporating the semantic, pixel-level, and scale information into a unified framework, PCRLv2 outperforms various SSL baselines in different labeling ratios significantly. It surpasses the previous conference version by clear margins, i.e., PCRLv1. Particularly, PCRLv2 seems to have more advantages in small labeling ratios. For instance, when the labeling ratio is 5\%, PCRLv2 outperforms PCRLv1 by 2.5 percents on average, which verifies the significance of multi-scale latent representations.
\begin{table*}[!htp]
    \centering
    \caption{Transfer learning on brain tumor segmentation (on BraTS). \textbf{WT}, \textbf{TC}, and \textbf{ET} stand for the whole tumor, tumor core, and enhancing tumor. For all SSL approaches, we use LUNA for pre-training, and then fine-tune the pre-trained model on BraTS with varying amounts of labeled data. Best results are bolded.}
    \resizebox{1.0\textwidth}{!}{
    \begin{tabular}{|c|c|ccc|c|ccc|c|ccc|c|ccc|c|ccc|}
    \hline
         \multirow{2}{*}{Methodology}&\multicolumn{4}{c|}{10\%}&\multicolumn{4}{c|}{20\%}&\multicolumn{4}{c|}{30\%}&\multicolumn{4}{c|}{40\%}&\multicolumn{4}{c|}{100\%} \\ \cline{2-21}
         & Mean & WT & TC & ET & Mean & WT & TC & ET & Mean & WT & TC & ET & Mean & WT & TC & ET & Mean & WT & TC & ET \\ 
         \hline
         \hline
         TS & 66.6 & 71.2 & 66.7 & 62.1 & 72.7 & 78.5 & 74.3 & 65.5& 76.7 &81.8& 77.9& 70.6& 77.1 & 82.3 & 78.3 & 70.9 & 81.5 & 86.8& 82.8&75.1\\
         MG~\cite{zhou2021models}&69.6 & 72.4& 71.4& 65.1 & 75.5 & 80.4 &77.3& 68.9 & 79.6 & 84.2& 80.6& 74.1 & 80.4 & 85.3 & 82.0& 74.0 & 82.4 & 87.1& 83.6& 76.6\\
         TransVW~\cite{haghighi2021transferable}&70.3 & 74.6 &71.7 &64.6 & 75.6 &79.9& 75.4& 71.5 & 79.1 & 83.8& 79.9& 73.6& 80.8 & 85.8 &82.1& 74.5 & 82.3 & 87.1 & 83.3&76.5\\
         Cube++~\cite{tao2020revisiting}&69.0 & 74.5& 70.6& 61.9 & 74.9 & 80.7& 75.9& 68.1 & 79.3 & 84.0  &79.4& 74.5 & 79.7 & 84.5& 80.0  &74.6& 82.2 & 87.2 & 82.4&77.0\\
         3D-CPC~\cite{taleb20203d}&70.1 & 76.7 &70.5 &63.1 &75.9 & 81.6 &75.6 &70.5 &79.4 & 84.6 &79.9 &73.7 &81.2 & 86.5 &81.8 &75.3 &82.9 &88.0 &83.3 &77.4\\
         \hline
         PCRLv1~\cite{zhou2021preservational} &71.6 & 76.9 &73.1& 65.2 & 77.6 & 81.4 &79.1 &72.7 & 81.1 & 84.9& 82.2& 76.6 & 83.3 & 87.5 &\textbf{84.6}& 78.2 & 85.0 & 89.0& \textbf{86.2}&80.2\\
         PCRLv2 & \textbf{73.0} & \textbf{77.7} &\textbf{74.3} &\textbf{67.2} & \textbf{78.8} & \textbf{83.2} & \textbf{79.4} & \textbf{74.0} & \textbf{82.1} & \textbf{85.1}& \textbf{82.7}& \textbf{78.7} & \textbf{84.1} & \textbf{87.9}& {84.5}& \textbf{80.1}& \textbf{85.6} & \textbf{89.4} & {85.9}& \textbf{81.7}\\
    \hline
    \end{tabular}}
    \label{tl:brats}
\end{table*}

\begin{table}[!htp]
    \centering
    \caption{Transfer learning on abdominal organ segmentation (on LiTS). We use LUNA for pre-training, and fine-tune the pre-trained model on LiTS with different amounts of labeled data. Best results are bolded.}
    \resizebox{0.9\columnwidth}{!}{
    \begin{tabular}{|c|c|c|c|c|c|}
    \hline
         \multirow{2}{*}{Methodology} & \multicolumn{5}{c|}{Labeling ratio}\\ \cline{2-6}
         & 10\% & 20\% & 30\% & 40\% & 100\%  \\
         \hline
         \hline
         TS & 71.1 & 77.2 & 84.1 & 87.3 & 90.7\\
         MG~\cite{zhou2021models} & 73.3 & 79.5 & 84.3 & 87.9 & 91.3\\
         TransVW~\cite{haghighi2021transferable} &73.8& 79.3& 85.5& 88.2 &91.4\\
         Cube++~\cite{tao2020revisiting} & 74.2& 79.3& 84.5& 88.2& 91.8\\
         3D-CPC~\cite{taleb20203d}&74.8& 80.2& 85.6& 88.9& 91.9\\
         \hline
         PCRLv1~\cite{zhou2021preservational} & 77.3 & 83.5 & 87.8 & 90.1 & 93.7\\
         PCRLv2 & \textbf{79.0} & \textbf{86.5}& \textbf{89.3}& \textbf{90.9}&\textbf{94.5}\\
    \hline
    \end{tabular}}
    \label{tl:lits}
\end{table}

\subsection{Semi-supervised pulmonary nodule detection}
In Table~\ref{ssl:luna}, we report the experimental results of semi-supervised pulmonary nodule detection. Interestingly, we observe narrowed performance gaps between TS and SSL baselines than those reported in Table~\ref{ssl:chest14}. One possible explanation is that the task of detecting pulmonary nodules is less sensitive to the amount of labeled data. Among all SSL baselines, Cube++ gives better performance when utilizing small amounts of labeled data, while 3D-CPC is more advantageous in large labeling ratios. In addition, we see TransVW quickly catching up with MG and Cube++ as the labeling ratio increases.

PCRLv1 outperforms previous SSL approaches in different labeling ratios by large margins. After incorporating multi-scale latent representations, PCRLv2 consistently surpasses PCRLv1 in a range of labeling ratios. When the baseline SSL methods show similar performance as the labeling ratio increases, PCRLv2 can still provide impressive improvements over PCRLv1 and previous SSL approaches.
\subsection{Transfer learning on chest pathology identification}
In Table~\ref{tl:chest14}, we validate the transferable ability of visual representations provided by different pre-training methodologies. Specifically, we compare PCRLv2 against train from scratch, ImageNet-based pre-training (IN), different SSL baselines, and PCRLv1.

Comparing MG/TransVW with IN, we see context restoration based SSL maintains the limited transferable ability. This phenomenon becomes more apparent when the target domain has quite limited annotations. The underlying reason is that semantic information plays a crucial role in transfer learning. In contrast, the significant performance gains brought by C2L and SimSiam again verify the effectiveness of comparative SSL. C2L and SimSiam still cannot outperform IN by significant margins, especially when considering that IN is more advantageous when the labeling ratio is 10\%. 

After integrating the benefits of context restoration based and comparative SSL, PCRLv1 is already capable of outperforming previous SSL methodologies by observable margins. Furthermore, by exploiting multi-scale semantic and pixel-level information, PCRLv2 achieves consistent improvements over PCRLv1 in overall and class-specific results in different labeling ratios. 

\subsection{Transfer learning on brain tumor segmentation}
We report the experimental results of applying transfer learning to brain tumor segmentation in Table~\ref{tl:brats}, where we use LUNA dataset for self-supervised pre-training and fine-tune the pre-trained model with different amounts of labeled data.

\begin{figure*}[htp]
    \centering
    \includegraphics[width=0.95\textwidth]{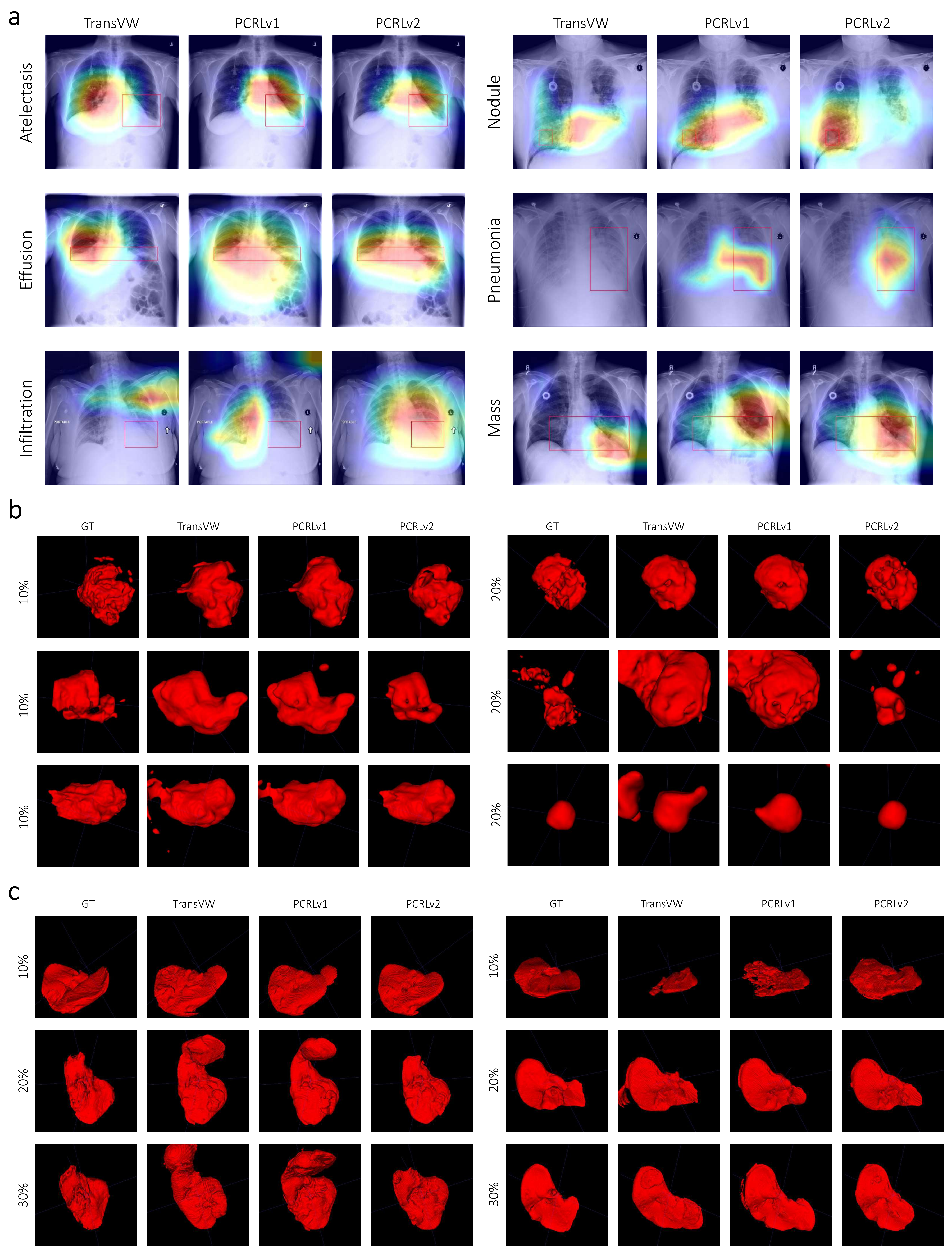}
    \caption{Visual interpretation of the transfer learning on chest pathology identification (\textbf{a}), and segmentation results of brain tumor (\textbf{b}) and liver (\textbf{c}). We mainly compare PCRLv2 against PCRLv1 and TransVW. Red boxes in the top figure \textbf{a} denote the ground-truth (GT) annotations from radiologists. In figure \textbf{b}, we present the segmentation results of the enhancing tumor (ET) from BraTS when the labeling ratios are 10\% and 20\%. Similarly in the bottom figure, we display the liver segmentation results in three different labeling ratios (10\%, 20\%, and 30\%).}
    \label{vis}
\end{figure*}

\begin{figure}[htp]
\centering
\includegraphics[width=1.0\columnwidth]{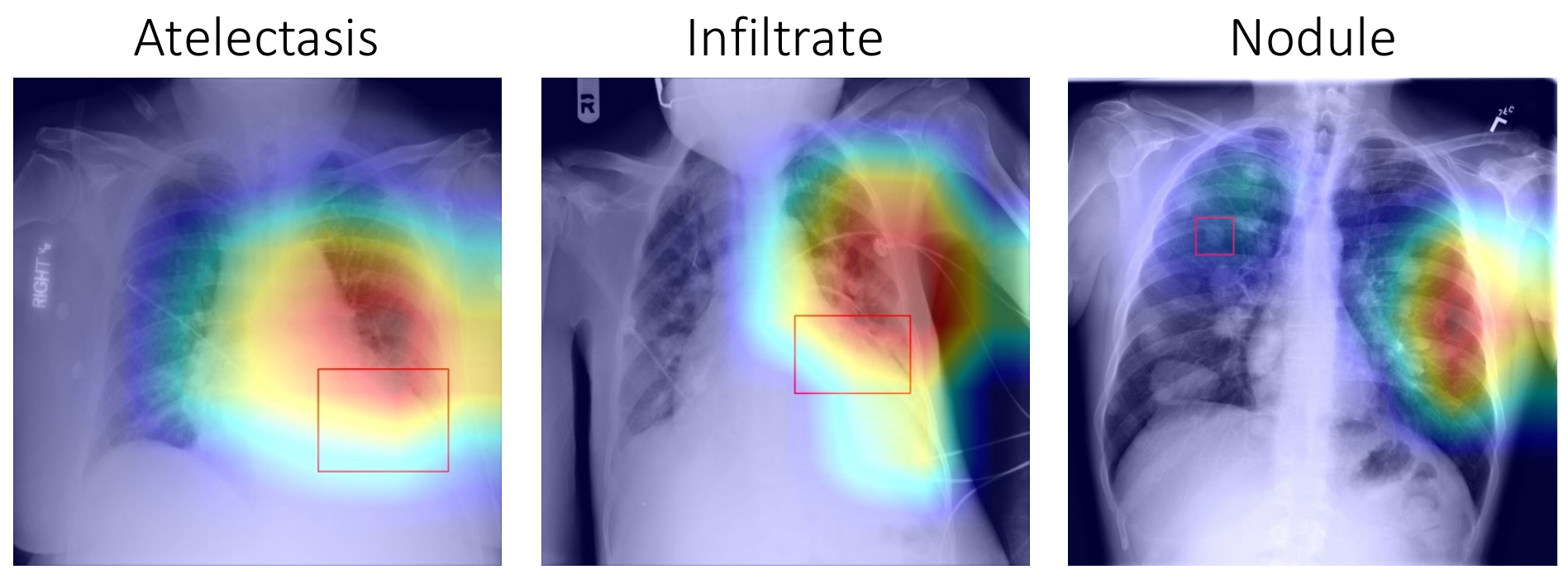}
\caption{Failure case analysis on chest pathology identification. Red boxes stand for the lesion areas delineated by radiologists. Images are from NIH ChestX-ray.}
\label{failures}
\end{figure}

Somewhat surprisingly, we find 3D-CPC does not outperform context restoration based SSL (MG, TransVW, and Cube++) as obviously as those in Tables~\ref{ssl:chest14},~\ref{ssl:luna}, and~\ref{tl:brats}. This comparison is consistent with our intuition: pixel-level information matters a lot in medical image segmentation. Again, PCRLv1 and PCRLv2 outperform previous SSL methodologies in all three classes by large margins. Compared to PCRLv1, PCRLv2 is more advantageous in segmenting the enhancing tumor (ET) regions, which are often smaller than WT and TC, and thus harder to segment. The performance gains on ET again verify the effectiveness of multi-scale latent representations, which advances the segmentation of small objects.

\subsection{Transfer learning on liver segmentation}
In Table~\ref{tl:lits}, we present the results of liver segmentation. There exist three observable phenomena. First, we see that all SSL approaches provide substantial performance gains over train from scratch. Second, we find the comparative methodology, i.e., 3D-CPC, achieves comparable segmentation performance to traditional context restoration based SSL. This phenomenon verifies the necessity of utilizing pixel-level information in medical image segmentation (similar results also appear in Table~\ref{tl:brats}). Last but not the least, PCRLv2 consistently outperforms PCRLv1 in all labeling ratios, which again validates the effectiveness of introducing multiple scales into SSL.

\subsection{Visual analysis}
In Fig.~\ref{vis}, we visually analyze the experimental results of transfer learning with limited annotations on chest pathology identification (10\%), brain tumor segmentation (10\% and 20\%), and liver segmentation (10\%, 20\%, and 30\%). Here, we compare PCRLv2 against generic SSL methodologies. Considering TransVW was developed on top of MG, we exclude MG and compare PCRLv2 against PCRLv1 and TransVW.

Fig.~\ref{vis}a presents the visual interpretation of chest pathology diagnoses using CAM~\cite{cam} on six different pathologies. We find that TransVW fails to capture the correct location of lesions on atelectasis, infiltration, nodule, and pneumonia. In comparison, PCRLv1 can generate more interpretable diagnosis results but still yields inconsistent predictions on infiltration and nodule. By integrating multi-scale latent representations, PCRLv2 can capture the small lesion areas on infiltration and nodule, resulting in centralized yet accurate diagnosis results.

In Fig.~\ref{vis}b and Fig.~\ref{vis}c, we visualize the segmentation results of the enhancing tumor (ET) on BraTS and liver on LiTS. Compared to TransVW and PCRLv1, PCRLv2 reduces the false positive predictions and contains richer fine-grained details. We believe such superiority of PCRLv2 can be attributed to the integration of multi-scale pixel-level and semantic information.

We also provide some failure examples in Fig.~\ref{failures}. One common characteristic of these detection results is that they include high-confidence predictions outside the lung area. However, in daily clinical practice, such anomalies should not be located outside the lung area. Similar phenomena have been reported in \cite{degrave2021ai}, where the authors summarized them as ``shortcuts'' that are common in learning systems based on neural networks. To mitigate this problem in self-supervised learning, we can add commonsense knowledge to pre-trained models. Besides, it is also necessary to develop more powerful machine learning tools for model interpretation in various downstream tasks.

\section{Conclusion}
We present a unified visual information preservation framework for self-supervised learning in medical imaging. This framework aims to encode the pixel-level, semantic, and scale information into latent representations simultaneously. To achieve this goal, we conduct multi-scale pixel restoration and feature comparison on the feature pyramid, which non-skip U-Net supports. The proposed PCRLv2 outperforms previous self-supervised pre-training approaches by large margins and yields consistent improvements over its conference version (PCRLv1) on four well-established datasets in both quantitative and qualitative validation. We will continue to explore how to optimally integrate different types of information into SSL in the future.

{\small
\bibliographystyle{ieee_fullname}
\bibliography{egbib}
}

\end{document}